\documentclass[lettersize,journal]{IEEEtran}
\usepackage{amsmath,amsfonts}
\usepackage{algorithmic}
\usepackage{algorithm}
\usepackage{array}
\usepackage{textcomp}
\usepackage{stfloats}
\usepackage{url}
\usepackage{verbatim}
\usepackage{graphicx}
\usepackage{cite}
\usepackage{subcaption}
\usepackage{multirow}
\usepackage{bbding}
\usepackage{xcolor}
\usepackage{placeins} 
\begin{document}

\title{Progressive Alignment Degradation Learning for Pansharpening}

\author{Enzhe Zhao, Zhichang Guo, Yao Li, Fanghui Song and Boying Wu
\thanks{Enzhe Zhao, Zhichang Guo, Yao Li, Fanghui Song and Boying Wu are with the Department of Computational Mathematics, School of Mathematics, Harbin Institute of Technology, Harbin 150001, China.}}




\maketitle

\begin{abstract}
Deep learning-based pansharpening \textcolor{black}{has been shown to effectively generate
high-resolution multispectral (HRMS) images. To create supervised ground-truth HRMS images, synthetic data generated using the Wald protocol is commonly employed. This protocol assumes that networks trained on artificial low-resolution data will perform equally well on high-resolution data. However, well-trained models typically exhibit a trade-off in performance between reduced-resolution and full-resolution datasets. } \textcolor{black}{In this paper, we delve into the Wald protocol and find that its inaccurate approximation of real-world degradation patterns limits the generalization of deep pansharpening models.} To address this issue, we propose the Progressive Alignment Degradation Module (PADM), which uses mutual iteration between two sub-networks, PAlignNet and PDegradeNet, to adaptively learn accurate degradation processes without relying on predefined operators. \textcolor{black}{Building on this, we introduce HFreqdiff, which embeds high-frequency details into a diffusion framework and incorporates CFB and BACM modules for frequency-selective detail extraction and precise reverse process learning.} These innovations enable effective integration of high-resolution panchromatic and multispectral images, significantly enhancing spatial sharpness and quality. Experiments and ablation studies demonstrate the proposed method's superior performance compared to state-of-the-art techniques. 
\end{abstract}


\section{Introduction}


\IEEEPARstart{R}{emote} sensing images with high spatial and spectral resolution are in high demand across various fields, including scene classification \cite{ref1,ref1_1}, semantic segmentation \cite{ref2, ref2_1}, and environmental monitoring \cite{ref3}. However, due to the physical limitations of current sensor technologies, data acquired by a single satellite sensor often fail to meet these high-quality standards. To address this challenge, pansharpening algorithms \cite{ref4, ref5} have been developed to fuse high spatial resolution panchromatic (PAN) images with low spatial resolution multispectral (LRMS) images, generating high-resolution multispectral (HRMS) images that effectively combine their respective advantages.

Traditional pansharpening methods primarily include component substitution (CS) \cite{ref_t1, ref2-1-1, ref2-1-2}, multi-resolution analysis (MRA) \cite{ref_t2, ref2-1-5, ref2-1-6}, Bayesian approaches \cite{ref9_1}, and variational optimization (VO) \cite{ref9_2}. These methods extract features from LRMS and PAN images using conventional tools and fuse them through manually designed rules to reconstruct HRMS images. However, their ability to capture deep relationships between LRMS and PAN images is limited, often resulting in constrained reconstruction quality. In recent years, deep learning has significantly advanced pansharpening by modeling complex nonlinear mappings \cite{ref10, ref11, ref13, ref14, ref15, ref16, ref16_1}. Supervised learning frameworks optimize the pansharpening process through end-to-end networks, enhancing spatial resolution and detail recovery while maintaining spectral consistency. Nevertheless, due to the lack of authentic high-resolution data, most methods rely on training data generated under Wald's protocol \cite{ref17}, where LRMS images are created by downsampling and blurring the original MS images, with the original MS images serving as ground truth. Although Wald's protocol provides paired training data, its assumption of a fixed degradation pattern may lead to mismatches between artificially degraded data and real low-resolution data, limiting performance in high-quality full-resolution fusion tasks. On the other hand, unsupervised models \cite{ref-unsuper-1, ref-unsuper-2, ref-unsuper-3} train directly on original MS data, avoiding Wald's protocol. However, these methods often rely on complex loss functions to optimize fusion quality and typically simplify the degradation process as linear, failing to capture the complex nonlinear degradation characteristics of real-world scenarios.

To accurately approximate the degradation process, we propose the Progressive Alignment Degradation Module (PADM), which dynamically captures potential degradation patterns by alternately training two sub-networks, PAlignNet and PDegradeNet. Unlike Wald's protocol, which relies on manually predefined degradation operators, PADM adopts a learning-driven approach to adapt to the complex and nonlinear degradation characteristics of real-world scenarios. This learning-driven degradation method also provides new insights for degradation modeling in image fusion tasks across other domains, offering broad applicability potential. Building on this, we introduce a diffusion framework to address the spatial and spectral fusion challenges in pansharpening. Diffusion models \cite{ref26, ref27, ref28}, known for their strong generalization capabilities on high-dimensional data, have emerged as a significant research direction in pansharpening \cite{ref32, ref33}. Theoretically, diffusion models can achieve polynomial-level small generalization errors, effectively overcoming the curse of dimensionality \cite{li2023generalization}. However, existing pansharpening diffusion models typically condition on PAN and LRMS images, embedding them into the diffusion architecture via CNNs, which often neglects the extraction of high-frequency spatial information due to limitations in the embedding structure. To address this, we propose the High-Frequency Guided Diffusion Model (HFreqDiff), which explicitly incorporates high-frequency information from PAN images as a condition, enabling selective frequency learning and spatial detail preservation. Combined with PADM, HFreqDiff achieves enhanced performance in both spectral fidelity and spatial resolution.

Our contributions can be summarized as follows:


- We propose a novel progressive alignment degradation module (PADM) that unsupervisedly learns degradation processes through iterative optimization, accurately modeling real-world degradation and minimizing spatial distribution discrepancies between simulated and real low-resolution data, enhancing adaptability to real-world conditions. 
 
- We design a high-frequency information learning module for the pansharpening diffusion framework, leveraging the low-frequency features of LRMS images to enhance high-frequency detail extraction from PAN images, improving the spatial fusion capability of the diffusion model.
 
- We present HFreqDiff, integrating the high-frequency learning module with Condition Fusion Blocks and Band-Aware Modulation Blocks to embed high-frequency details into the diffusion framework, enabling frequency-selective detail extraction and precise reverse-process learning. 

- Extensive experiments demonstrate HFreqDiff's superior quantitative and qualitative performance over state-of-the-art methods.  

The paper is organized as follows: Section 2 reviews related work, Section 3 details the proposed model, Section 4 provides experimental settings and analysis, and Section 5 concludes with future research directions.

\section{Related work}
\subsection{Traditional pansharpening method}
Traditional methods in pansharpening primarily include Component Substitution (CS), Multi-Resolution Analysis (MRA), Bayesian approaches, and Variational Optimization (VO). CS methods project the original Low-Resolution Multispectral (LRMS) image into a transformed domain, aiming to replace part or all of its spatial information with that of the Panchromatic (PAN) image. Common CS techniques include Intensity Hue Saturation (IHS) \cite{ref2-1-1}, Principal Component Analysis (PCA) \cite{ref2-1-2}, and Band-Dependent Spatial Detail (BDSD) \cite{ref2-1-3}. However, improper decomposition of spectral and spatial information can introduce artifacts. To address spectral distortion, MRA-based methods decompose the PAN image at multiple scales to extract spatial components, which are then injected into the LRMS image to enhance spatial details. The Modulation Transfer Function (MTF) and its variants are prominent examples of MRA techniques \cite{ref2-1-5, ref2-1-6}. Bayesian approaches aim to statistically model the relationship between the pansharpened result and the LRMS and PAN images. Since Fasbender et al. \cite{refb1} introduced Bayesian estimation theory, advancements have been made to tackle the ill-posed nature of this inverse problem \cite{refb2, refb3, refb4}. VO methods, a subset of Bayesian frameworks \cite{refb5}, formulate pansharpening as an optimization task by incorporating spatial and spectral consistency as regularization constraints. Ballester et al. \cite{ref2-1-7} pioneered the use of variational models for this task, while Liu et al. \cite{ref2-1-8} developed a method leveraging spatial fractional-order geometry and spectral-spatial low-rank priors to ensure spatial and spectral fidelity in the fused results. 


\subsection{Deep learning-based pansharpening methods}
In recent years, deep learning-based methods have rapidly advanced in pansharpening and become a research hotspot. According to whether ground truth or reference images are needed for training, these methods can be divided into two categories: supervised methods and unsupervised methods.


\subsubsection{Supervised methods}
Most supervised methods train networks at reduced resolution using the original MS image as ground truth, then apply the trained model directly to the original PAN and MS images to obtain full-resolution fused images. PNN \cite{ref10}, inspired by single-image super-resolution techniques, was the first CNN designed for pansharpening, achieving state-of-the-art performance at the time. PanNet \cite{ref11} explicitly injected high-frequency information from the PAN image into the upsampled LRMS image to preserve spectral information. Deng et al. \cite{ref19} proposed FusionNet, combining traditional methods with machine learning to enhance nonlinear modeling capabilities. To better balance spectral and spatial feature extraction, dual-stream networks \cite{ref-super-1,ref-super-2} were designed to learn more informative features. Additionally, leveraging Transformer's strength in global information aggregation, some CNN-based methods integrated Transformers to capture long-range dependencies \cite{ref2-2-3, ref2-2-4}.

Recently, in multi-modal image fusion, diffusion models \cite{ref26} have demonstrated superior generalization capabilities with full-resolution data. Dif-Fusion \cite{ref2-2-8} used the denoising network of diffusion models as a powerful feature extractor while decoupling the fusion process from the diffusion model. Building on this, PanDiff \cite{ref32} was the first to learn the data distribution of difference maps between HRMS and upsampled LRMS images from a conditional diffusion perspective. Zhou et al. \cite{ref33} further proposed a spatial-spectral integrated diffusion model, separately learning spatial and spectral features before fusion to improve feature discrimination and overall quality. Cao et al. \cite{ref35} introduced two conditional modulation modules within the diffusion framework to capture style and frequency information, reducing information loss.

\subsubsection{Unsupervised methods}
Since supervised training requires synthesizing reduced-resolution data, primarily following Wald's protocol to degrade the original data, the spatial degradation process in real-world scenarios is highly complex and difficult to accurately model with manually designed operators. This has led to the development of numerous unsupervised models that train directly on full-resolution images. For instance, based on GANs, \cite{ref-unsuper-1} proposed a dual-discriminator approach, with each discriminator dedicated to preserving spatial and spectral information. Xu et al. \cite{ref-unsuper-2} introduced adversarial learning between a generator and two discriminators, utilizing mean difference images to enhance the reconstruction quality of fusion results. Due to the absence of supervised information, some researchers have designed more complex loss functions to ensure the quality of fused images. Luo et al. \cite{ref-unsuper-3} employed structural similarity and no-reference quality metrics to guide network training, while Xing et al. \cite{ref34} proposed a self-supervised diffusion model for pansharpening, adopting a two-stage training process to pre-train a UNet and fine-tune the fusion head, demonstrating effective generalization on satellite datasets.

Although unsupervised methods reduce reliance on paired data and mitigate issues arising from resolution discrepancies, the lack of reference images significantly complicates the training process. These methods often depend on modeling the degradation process to compute loss functions, yet such estimations are frequently inaccurate.

\section{Proposed method}

\subsection{Motivation} 

Deep learning-based methods, particularly supervised ones, are built on Wald's protocol. The protocol defines a specific degradation process, where degraded data is generated by applying a Gaussian blur kernel, followed by downsampling of the PAN-MS pair. The original multispectral image is then used as the reference, with these degraded PAN-MS pairs serving as input samples for training the network. The degraded MS-PAN pairs can be formulated as
\begin{align}
	LRMS &= (MS_o\otimes k_m)\downarrow \\
	LRPAN &= (PAN_o\otimes k_p)\downarrow 
\end{align}
where $MS_o$ and LRMS represent the original MS image and its degraded version, respectively, and similarly for $PAN_o$ and LRPAN. $k_m$ and $k_p$ denote Gaussian blur kernels for the MS and PAN images. The operator $\otimes$ indicates the convolution operation and $ \downarrow$ represents the downsampling operation.

Under Wald's protocol, models are ideally expected to learn the anti-degradation operator during training on reduced-resolution datasets and generalize to full-resolution datasets. However, the effectiveness of this assumption heavily depends on the accuracy of the degradation modeling. If the degradation approximation is flawed, the anti-degradation operator learned by the model will be compromised, impairing its ability to generalize to high-resolution data.

\begin{figure}[htbp]
	\centering
	\includegraphics[width=0.48\textwidth]{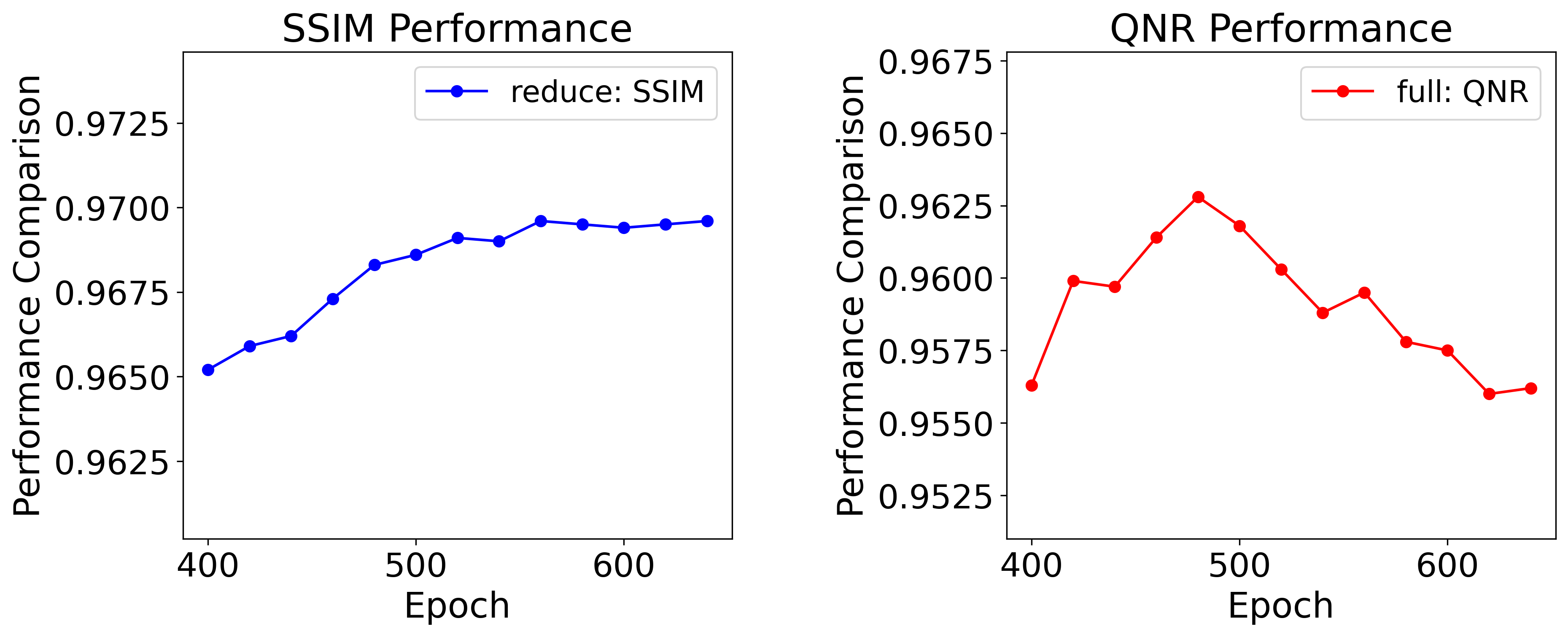}  
	\caption{The performance comparison of the reduced-resolution data and full-resolution data on WorldView-3 dataset. The left plot illustrates the SSIM metric for the reduced-resolution data, which generally shows a stable increasing trend as the number of training epochs increases. The right plot illustrates the QNR metric for the full-resolution data, where the QNR value decreases as the number of training epochs increases.}\label{fig:zhexiantu}
\end{figure}

Studies have shown that some models perform well on reduced-resolution data but poorly on full-resolution data \cite{ref-unsuper-3, ref-unsuper-4}, suggesting that the models may learn under incorrect degradation assumptions, limiting their generalization to high-resolution scenarios. Experimental results in Fig. \ref{fig:zhexiantu} validate this: when training the CNN model \cite{ref19} on the reduced-resolution WorldView-3 dataset, its performance improves with iterations, but on full-resolution data, both the visual quality of HRMS images and evaluation metrics significantly decline. This confirms that the inaccurate degradation assumptions in Wald's protocol restrict the model's performance on full-resolution data.

\subsection{Progressive Alignment Degradation Module}

To provide an accurate approximation of the degradation process, we propose an innovative progressive alignment downsampling module (PADM) that enables the network to dynamically learn the degradation patterns, rather than relying on \textcolor{black}{artificial} degradation \textcolor{black}{priors}. 

The key idea of PADM is to progressively approximate the degradation operation using both the PAN and MS images, based on the fact that the degraded PAN image should retain the same spatial information as the MS image. To achieve this, we design a neural network that takes the PAN image as input and produces the degraded PAN image as output to approximate the degradation operation. Similarly, another network takes the MS image as input and the degraded PAN as output to model the spatial information alignment operation. When both networks are carefully designed, their common output—the degraded PAN image—serves as an effective constraint, ensuring consistency between them. However, in the absence of a closed-loop loss, multiple solutions may exist for the degradation approximation. To address this, we employ an iterative scheme, initializing with a sufficiently accurate estimate of the degraded PAN image. Through iterative training, the networks progressively converge to an accurate degraded PAN image, ultimately yielding a model that effectively approximates the degradation operation.

As described above, the proposed PADM consists of two sub-networks: PDegradeNet (Progressive Degradation Network) and PAlignNet (Progressive Alignment Network), which are trained using an interactive iterative strategy. \textcolor{black}{PDegradeNet takes the PAN image as input to learn the degradation operation, while PAlignNet takes the MS image as input to approximate the spatial information alignment operation.} The degraded PAN image learned by the networks is an ill-posed problem with multiple solutions. Therefore, we incorporate properties of degradation and spatial alignment operations into the design of corresponding networks.

\subsubsection{Equivariant Convolution}

Let \( \mathcal S \) represent the function space of the degradation approximation, \( f \in \mathcal S \) represent a degradation network, \( x \) represent the original image, and \( x_d \) represent the degraded image of \( x \), i.e., \( f(x) = x_d \). Assuming there exists an ideal degradation network \( f^* \), then \( x_d^* \) represents the ideal degradation result, i.e., \( f^*(x) = x_d^* \). The ideal degradation network \( f^* \) should have the following properties: (1) it should possess equivariance, such as rotation invariance; (2) \( x_d^* \) should retain the low-frequency information of the original image without producing artifacts or high-frequency noise; (3) \( x_d^* \) should be smooth, i.e., blurred relative to the original image.

To achieve the desired properties of \( f^* \), we adapt the approach of parameterized filtering. The basic idea of the parameterized filtering is to define the objective function as the linear combination of a set of basis functions $\{\psi_n\}_{n=1}^N$, aiming to get a learnable functional filter $\psi : \mathbb{R}^2 \rightarrow \mathbb{R}$. Formally, it can be expressed as
\begin{equation}\label{eq:filter}
	\psi(v) = \sum_{n=1}^N w_n \psi_n(v)
\end{equation}
where $v \in \mathbb{R}^2$  denotes the 2D spatial coordinates, $N$ is the number of basis functions, and $w_n$ is the $n$-th coefficient of the learnable parameter $w$.

Currently, various types of basis functions have been used for filter parameterization \cite{ref25_3, ref25_4}. \textcolor{black}{To better meet the desired properties of $f^*$,} we introduce a Fourier series expansion-based filter parameterization method. \textcolor{black}{The Fourier series expansion has a natural advantage in directly handling the frequency components of the signal, making it consistent with the degradation process.}


\textbf{2D Fourier bases:} A \( p \times p \) discrete filter \( \tilde{\phi} \) can be interpreted as the discretized version of an underlying 2D function \( \phi(v) \), which is uniformly sampled over the area \( [-(p-1)h/2, (p-1)h/2]^2 \) in \(\mathbb{R}^2 \), where \( h \) denotes the mesh size of the image. In formal terms, the 2D Fourier bases are given by:
\begin{align}
	\phi^c_{kl}(v) &= \Omega(v) \cos \left( \frac{2\pi}{ph} [k, l] \cdot \begin{pmatrix} v_1 \\ v_2 \end{pmatrix} \right)\\ 
    \phi^s_{kl}(v) &= \Omega(v) \sin \left( \frac{2\pi}{ph} [k, l] \cdot \begin{pmatrix} v_1 \\ v_2 \end{pmatrix} \right)
\end{align}
where \( k, l = 0, 1, \cdots, p - 1 \), and \( \Omega(x) \geq 0 \) is a radial mask function that satisfies \( \Omega(x) = 0 \) when \( \| v \|_2 \geq (p + 1/2)h \). Subsequently, the real part of the 2D inverse DFT for \( \tilde{\varphi} \) (with a circular mask) can be written in the following Fourier series expansion form:
\begin{equation} \label{eq:DFT}
	\phi(v) = \sum_{k=0}^{p-1} \sum_{l=1}^{p-1} (a_{kl} \phi^c_{kl}(v) + b_{kl} \phi^s_{kl}(v))
\end{equation}
where \( a_{kl} \) and \( b_{kl} \) are the expansion coefficients. Eq. \ref{eq:DFT} is a particular case of Eq. \ref{eq:filter}, indicating that it belongs to a filter parameterization method.

\textbf{F-conv:} Instead of the classic 2D Fourier bases, we adapt the Fourier series expansion-based equivariant convolution (F-conv) \cite{ref25_4}, which is an improved parameterization strategy for 2D filters based on Fourier series expansion. Formally, the bases of F-conv can be expressed as follows
\begin{align}
	\varphi^c_{kl}(v) &= \Omega(v) \cos \left( \frac{2\pi}{ph} ( k - \left\lfloor \frac{p}{2} \right\rfloor, l - \left\lfloor \frac{p}{2} \right\rfloor) \cdot \begin{pmatrix} v_1 \\ v_2 \end{pmatrix} \right) \\
    \varphi^s_{kl}(v) &= \Omega(v) \sin \left( \frac{2\pi}{ph} ( k - \left\lfloor \frac{p}{2} \right\rfloor,l - \left\lfloor \frac{p}{2} \right\rfloor) \cdot \begin{pmatrix} v_1 \\ v_2 \end{pmatrix} \right) 
\end{align}
where $k, l = 0, 1, \cdots, p-1$ and $\Omega(x) \geq 0$ is the aforementioned radial mask function. The filter parametrization is
\begin{equation} \label{eq:fconv}
	\varphi(v) = \sum_{k=0}^{p-1} \sum_{l=1}^{p-1} (a_{kl} \varphi^c_{kl}(v) + b_{kl} \varphi^s_{kl}(v))
\end{equation}
F-conv has been proven to have exact equivariance in the continuous domain, with only approximation errors introduced during discretization\cite{ref25_4}, \textcolor{black}{thus better preserving the equivariance. That is, shifting and rotating an input image of F-conv is equivalent to shifting and rotating all of its intermediate feature maps and the output image. In other words, the translation and rotation symmetries are preserved throughout the F-conv layers which is the necessary property of the degradation operation. Moreover, F-conv approximates the degradation operation in the frequency domain, which is consistent with the goal of smoothing the original image from a signal processing perspective.}

\begin{figure}[htbp]
	\centering
	\includegraphics[width=0.48\textwidth]{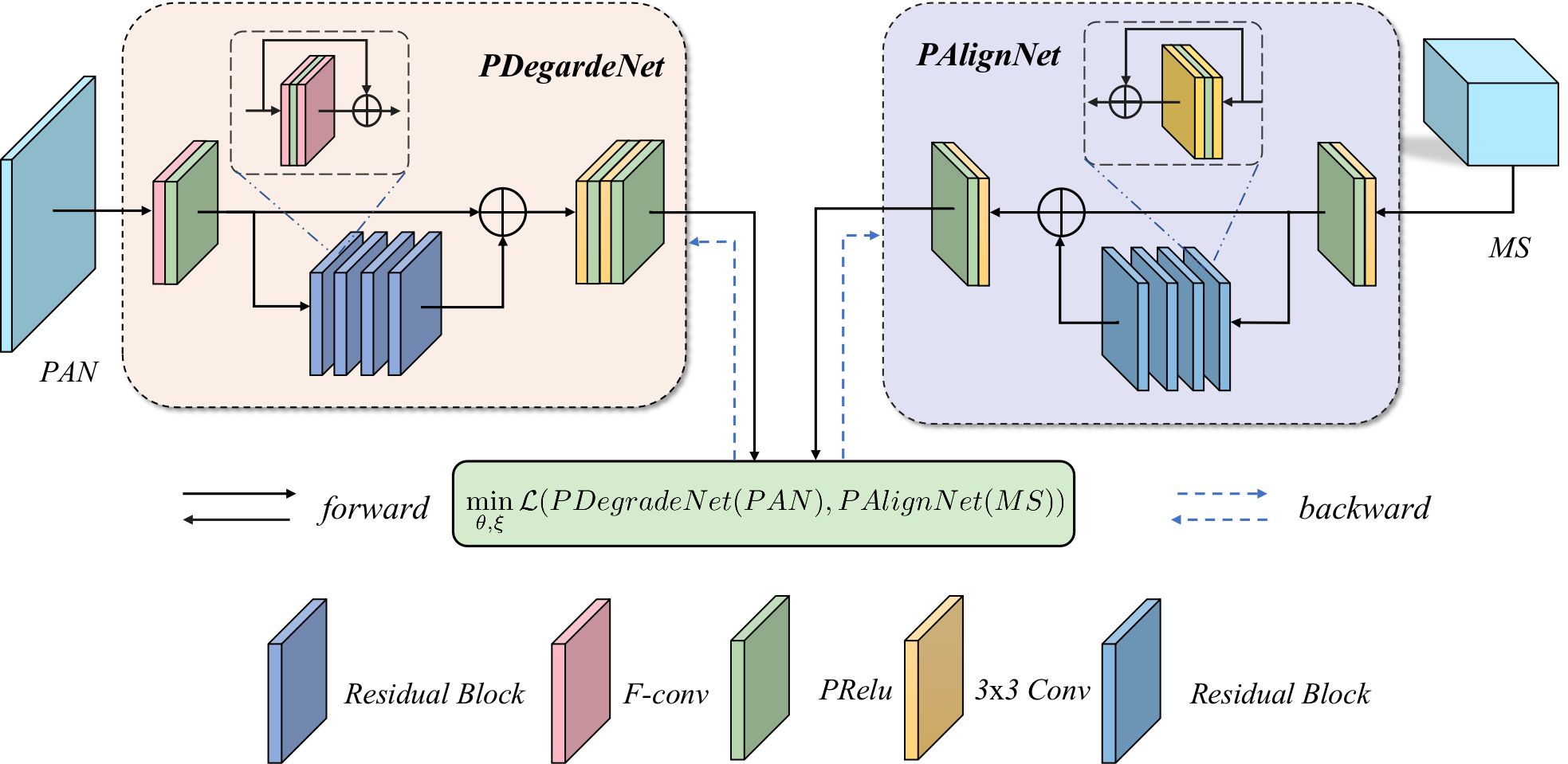}
	\caption{Network structure of PADM. PAlignNet takes the original MS image as input and performs an initial alignment with $LRPAN$. Subsequently, PDegradeNet uses the original PAN image as input and employs the output of PAlignNet as an intermediate result to further learn the degradation operation.}\label{fig:padm}
\end{figure}

\subsubsection{PAlignNet}
The network architecture of PAlignNet, as illustrated in Fig. \ref{fig:padm}, is specifically designed for spatial information alignment between low-resolution PAN and MS images. The lightweight network architecture comprises three main components: an initial $3\times3$ convolutional layer for feature extraction, followed by multiple residual blocks for deep feature representation, and a final $3\times3$ convolutional layer for feature reconstruction. Each residual block employs $3\times3$ convolutional kernels and incorporates residual connections to facilitate feature propagation and utilization. \textcolor{black}{The network utilizes shallow $3\times3$ convolutional layers to ensure a small spatial effective receptive field, thereby constraining the spatial information of the output to the local spatial information of the input. This design allows the network to focus on local spectral information for decolorization rather than relying on spatial information.}

\subsubsection{PDegradeNet} The PDegradeNet is designed to learn an adaptive degradation operation, as shown in Fig. \ref{fig:padm}. Its architecture is similar to that of PAlignNet. The network begins with an F-conv layer for initial feature extraction, followed by multiple cascaded residual modules constructed using F-conv layers for deep feature learning. Finally, the network employs three successive $3\times3$ convolutional layers to adjust the channel dimensions of the features. The adoption of F-conv in the network design ensures feature equivariance and allows the network to handle degradation directly in the frequency domain.

\subsubsection{Iterative Alternating Training}

The proposed PAlignNet and PDegradeNet are trained using a novel iterative alternating scheme, allowing the network to adaptively learn degradation patterns based on the spatial distribution of PAN and MS images. Specifically, the process begins with the initialization of a degraded PAN image $LRPAN \in \mathbb{R}^{\frac{H}{r}\times \frac{W}{r}\times C}$, where $r$ is the downsampling scale. PAlignNet takes the original MS image as input and performs an initial alignment with $LRPAN$. During this process, PAlignNet learns to decolorize the MS image based on the spatial information of $LRPAN$. Subsequently, PDegradeNet uses the original PAN image as input and employs the output of PAlignNet as an intermediate result to further learn the degradation operation. PAlignNet is then trained again based on the output of PDegradeNet, and this iterative process is repeated. The two networks interacting and refining each other's outputs until convergence is achieved, resulting in the final degradation network.

\textcolor{black}{
The process can be mathematically expressed as
\begin{align}
    \hat{\theta}^{(i)} = \mathop{\arg\min}\limits_{\theta^{(i)}}\;\mathcal{L}_{align}(&\text{PAlignNet}(MS;\theta^{(i)}), \notag\\
    &\text{PDegradeNet}(PAN;\hat{\xi}^{(i-1)})) \label{eq:fnet}\\
    \hat{\xi}^{(i)} = \mathop{\arg\min}\limits_{\xi^{(i)}}\;\mathcal{L}_{degrade}(&\text{PAlignNet}(MS;\hat{\theta}^{(i)}), \label{eq:gnet}\notag\\
    &\text{PDegradeNet}(PAN;\xi^{(i)}))
\end{align}}
where $\theta$ and $\xi$ represent the parameters of PAlignNet and PDegradeNet, respectively. 
The index $i=1,\hdots, I$ denotes the current iteration. In our experiments, we set the total number of iterations $I\geq 2$ to ensure sufficient training.
The initial value of the iterative training is $\text{PDegradeNet}(PAN;\hat{\xi}^{(0)})=\text{Downsample}(PAN)$. The loss function of PAlignNet is defined as: \
\begin{align} \label{eq:loss_align}
	\mathcal{L}_{align} = 
    \frac{1}{N} \sum_{n=1}^{N} \| \text{PAlignNet}(MS_n;\theta) - LRPAN \|_F^2
\end{align}
where \(N\) represents the total number of samples, and \(\| \cdot \|_F\) denotes the Frobenius norm. The loss function of PDegradeNet is similarly defined as:
\begin{align}\label{eq:loss_degrade}
	\mathcal{L}_{degrade} = 
    \frac{1}{N} \sum_{n=1}^{N} \| \text{PDegradeNet}(PAN_n;\xi) - LRPAN \|_F^2
\end{align}

In the $i$-th iteration, PAlignNet is optimized using the fixed parameters of PDegradeNet from the previous iteration. Subsequently, PDegradeNet is optimized using the fixed parameters of PAlignNet from the current iteration's previous step. The network parameters of PAlignNet and PDegradeNet can be optimized in an alternating manner. 



PADM provides an unsupervised framework to approximate the degradation operation. It assumes that the optimal degradation approximation is close to the initial guess, i.e., the downsampling, and can be achieved through iterative alternating optimization of the two networks. Through PADM, the degradation operation can be learned from real data rather than relying on preset fixed patterns, i.e., Gaussian blurs and downsampling. This enables the model to better handle unknown degradation processes encountered in real-world scenarios.

\subsection{High-Frequency-Guided Diffusion Model (HFreqDiff)}

Building on PADM, we introduce HFreqDiff, which embeds high-frequency details into a diffusion framework while incorporating Condition Fusion Blocks (CFB) and Band-Aware Modulation Blocks (BAMB) to enable frequency-selective learning and detail preservation.

\subsubsection{High-Frequency Detail Learning Module (HDLM)}

\begin{figure}[htbp]
	\centering
	\includegraphics[width=0.48\textwidth]{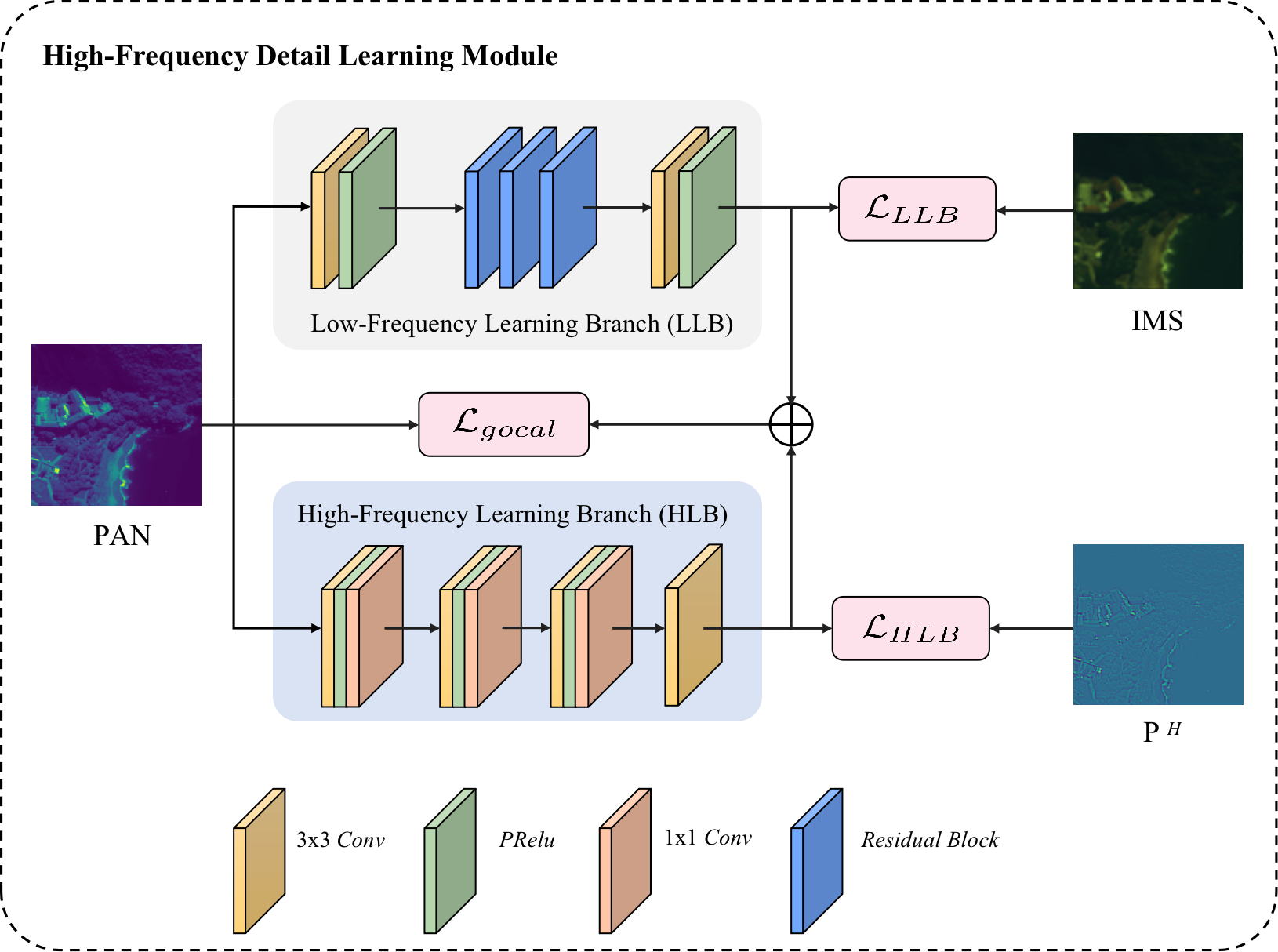}
	\caption{The architecture of HDLM.}\label{fig:hdlm}
\end{figure}

To enhance high-frequency information guidance for the diffusion model, we propose a high-frequency detail learning module designed to recover the high-frequency components missing in the MS image. The structure of this module, shown in Fig. \ref{fig:hdlm}, consists of two branches: a high-frequency learning branch (HLB) and a low-frequency learning branch (LLB). Specifically, we generate a high-frequency prior \( P^H \) of the PAN image using \( P^H = PAN - P^L \), where \( P^L \) is obtained by applying a low-pass filter to \( PAN \). The HLB utilizes \( P^H \) as a reference to extract high-frequency components from \( PAN \), including image details, edges, and textures. The loss function of the HLB is
\begin{equation}
	\mathcal{L}_{HLB} =  \frac{1}{N} \sum_{n=1}^{N} \left\| \text{HLB} \left( PAN_n \right)-  P^H_n  \right\|_F^2
\end{equation}
On the other hand, the LLB uses $MS$ as a reference to extract the corresponding low-frequency features, with the loss function defined as follows:
\begin{equation}
	\mathcal{L}_{LLB} =  \frac{1}{N} \sum_{n=1}^{N} \left\| \text{LLB} \left( PAN_n \right)- IMS_n  \right\|_F^2
\end{equation}
where $IMS_n=\text{Interpolate}(MS_n)$ is the interpolated MS image.

Additionally, to ensure spatial consistency between the high-frequency and low-frequency features, we fuse the outputs of both branches and align the resulting feature map with the original PAN image. The loss function can be expressed as:
\begin{align}
    &\mathcal{L}_{glocal} = \notag\\
    &\frac{1}{N} \sum_{n=1}^{N} \left\| \text{HLB} \left(PAN_n \right) + \text{LLB} \left( PAN_n \right) - PAN_n \right\|_F^2
\end{align}
As a result, the combination of three losses constitutes the total loss for the module, which can be represented as
\begin{equation}
	\mathcal{L}_{tocal} = \mathcal{L}_{HLB}  + \lambda_1 \mathcal{L}_{LLB} +\lambda_2 \mathcal{L}_{glocal} 
\end{equation}
Where $\lambda_1, \lambda_2$ are the weight hyperparameters. Finally, we use the output of the trained HLB, i.e. $P^h=\text{HLB}(PAN)$ as the one of the conditional input for HFreqDiff to provide additional high-frequency information.

\begin{figure}[htbp]
	\centering
	\begin{minipage}[b]{0.15\textwidth}
		\centering
		\includegraphics[width=\textwidth]{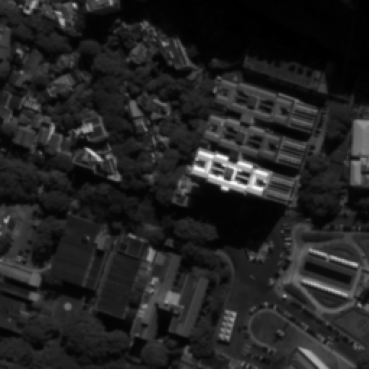}
	\end{minipage}%
	\hspace{0.01\textwidth} 
	\begin{minipage}[b]{0.15\textwidth}
		\centering
		\includegraphics[width=\textwidth]{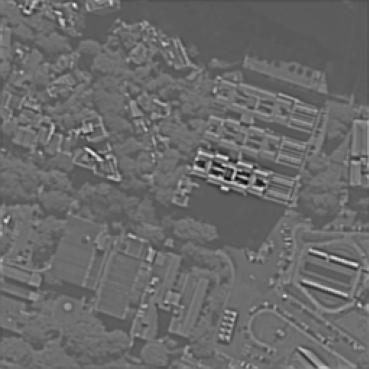}
	\end{minipage}%
	\hspace{0.01\textwidth} 
	\begin{minipage}[b]{0.15\textwidth}
		\centering
		\includegraphics[width=\textwidth]{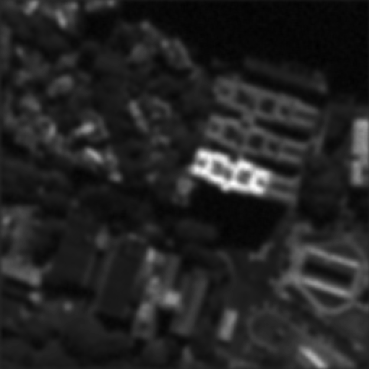}
	\end{minipage}
	
	\caption{Visual presentations of feature maps extracted by HDLM. From left to right: the PAN image, the high-frequency features output by the HLB, and the low-frequency features output by the LLB.} \label{fig:pan_hplp}
	
\end{figure}

To verify the effectiveness of the HLB in learning to separate key high-frequency features from the PAN image with support from the LLB, we present visualizations of the feature maps extracted by both HLB and LLB, as shown in Fig. \ref{fig:pan_hplp}. From the figure, it can be observed that the feature maps extracted by the HLB effectively preserve the edge details and fine structures of the image.

\subsubsection{Architecture of the Conditional Noise Predictor}

\begin{figure*}[htbp]
	\centering
	\includegraphics[width=0.75\textwidth]{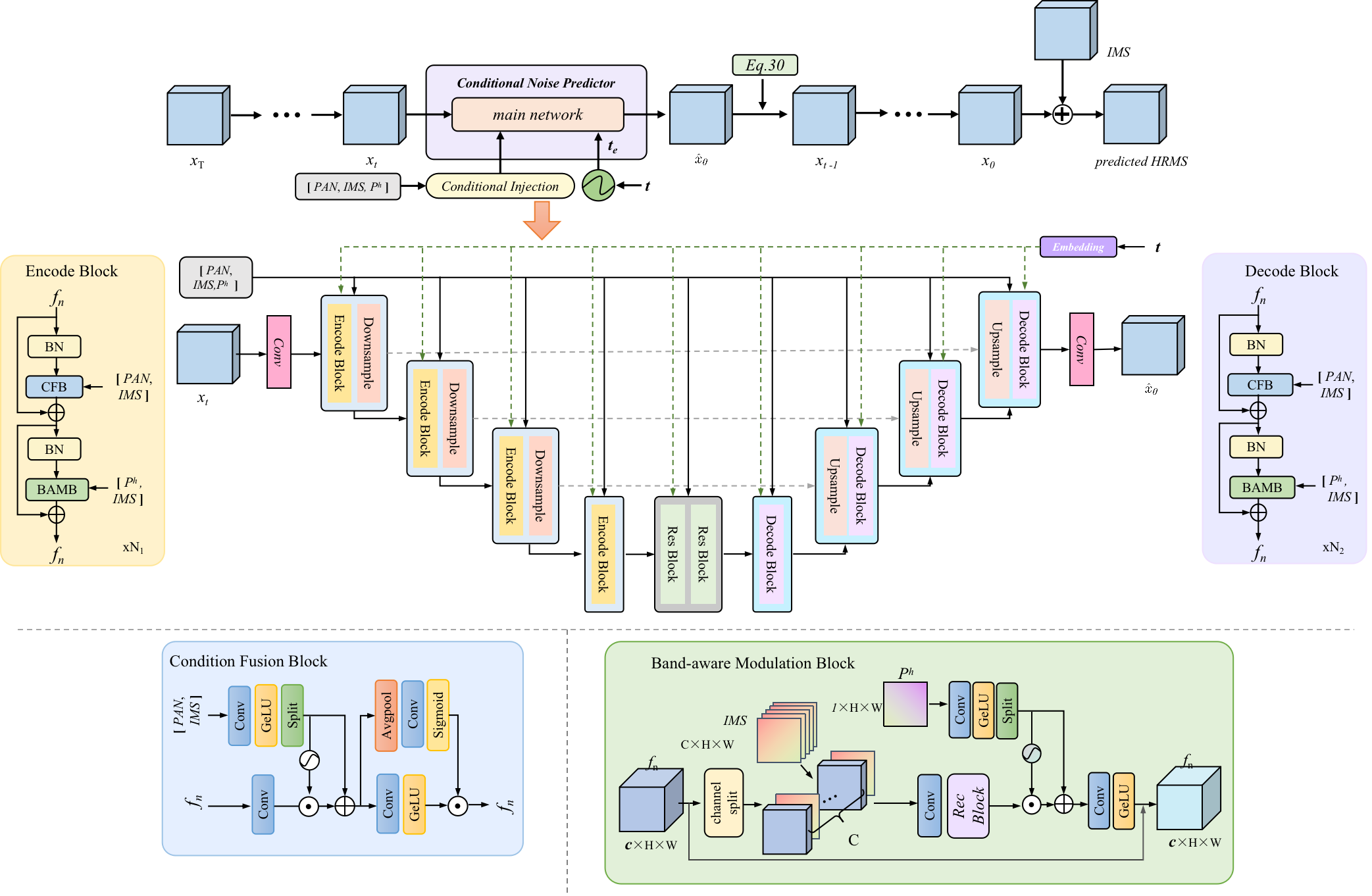}
	\caption{The inference procedure and the specific architecture of the conditional noise predictor in HFreqDiff are as follows. The conditional noise predictor adopts an encoder-decoder structure, which includes Condition Fusion Blocks (CFB) and Band-aware Modulation Blocks (BAMB).}\label{fig:aadiff}
\end{figure*}

The key high-frequency features learned through HDLM are utilized as the prediction condition for HFreqDiff, along with the PAN image and the interpolated MS image. The detailed architecture of the conditional noise predictor in HFreqDiff is shown in Fig. \ref{fig:aadiff}. The conditional noise predictor adopts an encoder-decoder structure, where each stage is specifically designed to capture and refine critical spatial and frequency features.

Given $x_t\in \mathbb{R}^{C\times H \times W}$,  we extract shallow features $f_n \in \mathbb{R}^{c\times h \times w}$ using convolution. Subsequently, $f_n$ passes through a four-stage encoder-decoder network for deep feature extraction. Each stage consists of a stack of varying numbers of encoding or decoding blocks, designed to enhance frequency details by fully utilizing complementary information.
As shown in Fig. \ref{fig:aadiff}, the encoding and decoding blocks consist of Condition Fusion Blocks (CFB), which take $PAN$ and $IMS$ as conditions to fuse spatial and spectral features from the PAN and MS images. Then, Band-aware Modulation Blocks (BAMB) use HDLM and the key high-frequency features $P^h$ extracted from HDLM as conditions to further enrich the features for local details and texture structures.   

\textbf{Condition Fusion Block (CFB):} CFB is designed to extract information representations from the PAN and LRMS images to establish a rough structural representation. The feature maps extracted via convolution from $PAN$ and $IMS$ are used as dynamic modulation parameters \cite{refm1, refm11}, which apply radiometric transformation to each intermediate feature map in the U-Net, adaptively influencing the output. Mathematically, the procedure can be expressed as
\begin{align}
	\alpha_n, \beta_n &= \text{Split}(\text{GeLU}(\text{Conv}([PAN, IMS])))\\
	\hat{f_n} &= \text{Sigmoid}(\alpha_n) \odot \text{Conv}(f_n) + \beta_n,\\
	f_n &= \text{GeLU}(\text{Conv}(\hat{f}_n))\odot \text{Sigmoid}(\text{Conv}(\text{Avgpool}(\hat f_n)))
\end{align}
where$f_n\in \mathbb{R}^{C\times H \times W}$ denotes input and output feature for the $n$-th encoder or decoder, $\odot$ denotes element-wise multiplication. The Conv consists of a $1 \times 1$ convolution and a $3 \times 3$ depth-wise convolution layer.

\textbf{Band-aware Modulation Block (BAMB):} Each band of a MS image carries unique information. A thorough analysis of their inherent properties and effective utilization can significantly enhance network design. BAMB is specifically designed to fully exploit modality-aware and band-aware characteristics. 
First, we split the features $f_n$ into multiple groups along the channel dimension, i.e., $\text{Split}(f_n) = \{f_n^1, ..., f_n^C\}$, where the number of groups $C$ is set to match the number of channels in the MS image. Then, $IMS$ is introduced by concatenating it with the corresponding group features along the channel dimension.
We adapt the modulation operation to inject the high-frequency features 
$P^h$ into each band, selectively and explicitly enhancing its feature representation by incorporating the texture-rich PAN features.
Finally, the modulated feature is integrated through channel-wise concatenation and then added to the input features $f_n$ to produce the final output.

The whole process is formulated as

\begin{align}
	\{f_n^1, ..., f_n^C\} &= \text{Split}(f_n)\\ 
    \{m_n^1, ..., m_n^C\} &= \text{Split}(IMS)\\
	f_n^k &= \text{Conv}(\text{Concat}([f_n^k, m_n^k])), \; k=1, \hdots, C \\
	\gamma_k, \eta_k &= \text{Split}(\text{GeLU}(\text{Conv}(P^h))) \\
	\tilde{f}_n^k &= \text{Sigmoid}(\gamma_k) \odot f_n^k + \eta_k \\
	f_n &= \text{Concat}([\tilde{f}_n^1, \hdots, \tilde{f}_n^C]) + f_n
\end{align}

\subsection{Diffusion Process of HFreqDiff}
HFreqDiff is a diffusion model conditioned on the PAN image, MS image, and high-frequency features. 
Inspired by previous works \cite{ref32, ref35, ref36}, instead of directly predicting the high-resolution HRMS image, we employ a residual prediction approach to estimate the difference between $HRMS\in \mathbb{R}^{C \times H\times W}$ and the interpolated MS image, i.e., $IMS\in \mathbb{R}^{C\times H\times W}$. We denote the difference as the input residual image $x_0=HRMS-IMS$. 
This strategy effectively mitigates the difficulty of converting HRMS into Gaussian noise and performing regression reconstruction over a limited number of time steps, while preserving critical details and structural information. 
The inference and training processes are as follows:

\subsubsection{Sampling Process}


Starting from $x_0$, the forward diffusion process progressively adds Gaussian noises to obtain a series of images $\{x_1, ..., x_T\}$,until it converges to a Gaussian distribution, $x_T \sim \mathcal{N}(0, I)$,  as formulated by
\begin{equation}
	q(x_t | x_{t-1}) := \mathcal{N}(x_t; \sqrt{1-\beta_t} x_{t-1}, \beta_t I),
\end{equation}
where $\beta_t$ is a predefined variance schedule in time step $t \in [0, T]$. By setting $\alpha_t = 1 - \beta_t$ and $\bar{\alpha}_t = \prod_{s=1}^{t} \alpha_s$, the forward process allows sampling of $x_t$ at any arbitrary timestep $t$ in closed form

\begin{equation}
	q(x_t | x_0) = \mathcal{N}(x_t; \sqrt{\bar{\alpha}_t} x_0, (1 - \bar{\alpha}_t) I)
\end{equation}
Through the reparameterization trick, we can derive $x_t$ directly from $x_0$ as
\begin{equation}
x_t(x_0, \epsilon) =\sqrt{\bar{\alpha}_t} x_0 + \sqrt{1 - \bar{\alpha}_t} \epsilon, \; \epsilon \sim \mathcal{N}(0, I)
\end{equation}

\subsubsection{Training Process}
Based on conditional DDPM,  the reverse process recover $x_0$ from $x_T$ via iterative denoising using a conditional noise predictor $x_\theta$. To perform denoising at each timestep, $x_\theta$ is designed to predict the original $x_0$ rather than the added noise $\epsilon_t$, since this empirically is beneficial for Pansharpening.
This conditional noise predictor $x_\theta$ allows HFreqDiff to iteratively refine the noisy input, guided by the conditioned information $cond=(PAN, IMS, P^h)$, toward an accurate reconstruction of the $x_0$. The loss function $\mathcal{L}$ takes the following form,
\begin{equation}
	\mathcal{L} = \mathbb{E} \left[ \| x_0 - {x}_\theta ({x}_t, cond, t) \|_1 \right]
\end{equation}

We opted for the DDIM approach \cite{ref37} to accelerate the sampling process,  which performs each step of the sampling with the update rule
\begin{align} \label{eq:xt}
    {x}_{t-1} =& \sqrt{\bar{\alpha}_{t-1}} {x}_\theta ({x}_t, cond, t) + \notag\\
    &\sqrt{1 - \bar{\alpha}_{t-1} - \sigma_t^2} \epsilon_\theta ({x}_t, cond, t)+ \sigma_t \epsilon
\end{align}
where the term $\epsilon_\theta (x_t, cond, t)$ is derived from Tweedie’s formula \cite{ref38} as
\begin{equation} \label{eq:xt1}
	\epsilon_\theta ({x}_t,cond, t) = \frac{{x}_t - \sqrt{\bar{\alpha}_t} {x}_\theta ({x}_t, cond, t)}{\sqrt{1 - \bar{\alpha}_t}}
\end{equation}
In this formula, $\sigma_t = \sqrt{\frac{1 - \bar{\alpha}_{t-1}}{1 - \bar{\alpha}_t}} \cdot \sqrt{\frac{1 - \bar{\alpha}_t}{\bar{\alpha}_{t-1}}}$ corresponds to the DDPM sampling. When $\sigma_t = 0$, the sampling process becomes deterministic. By leveraging these sampling equations, the sampling speed can be accelerated, while maintaining both simplicity and outstanding pansharpening performance.

\section{Experiments}

This section details the implementation, including methodologies, technical aspects, and datasets. It also outlines performance benchmarks for evaluating the pansharpening model and presents experimental results and ablation studies to quantitatively validate the proposed method's effectiveness.

\subsection{Implementation details}

For all experiments, we establish a virtual Anaconda environment with Python 3.8.5 and PyTorch 1.10.0 as the standard. The specific graph computation platform contains CUDA 11.1, CUDNN 8.0.4, and TensorRT 7.2.3.4 on a NVIDIA GeForce RTX4090 GPU. We use AdamW optimizer with an initial learning rate of 1e-4 to minimize $\mathcal{L}_{simple}$. For the diffusion model, the initial number of model channels is 32, the diffusion time step used for training in the pansharpening is set to 500, while the diffusion time step for sampling is set to 25 according to Eqs. \ref{eq:xt} and \ref{eq:xt1}. The exponential moving average (EMA) ratio is set to 0.999. The total training iterations for the WV3 and QB datasets are set to 100k and 150k iterations, respectively. 

\subsection{Datasets}
To demonstrate the effectiveness of our HFreqDiff, we conducted experiments on widely used pansharpening datasets. The PanCollection dataset \cite{ref39}, which includes data from two satellites— WorldView-3 (8 bands) and QuickBird (4 bands), is used to thoroughly and fairly evaluate our PanDiff model against other state-of-the-art methods. These three datasets encompass various geographical regions, each with different bands and spatial resolutions. Specifically, the spatial resolution of PAN images from QuickBird is 0.6 m, while its corresponding MS images have resolutions of 2.4 m, each with four bands: red, green, blue, and near-infrared. In comparison, WorldView-3 offers higher spatial resolution, with PAN images at 0.3 m and MS images at 1.2 m, as well as an expanded set of bands that includes coastal, yellow, red edge, and near-infrared-2. To better evaluate performance, we perform the reduced-resolution and full-resolution experiments to compute the reference and non-reference metrics, respectively. 

\subsection{Benchmark}
To assess the performance of our HFreqDiff, we compare it with previous state-of-the-art pansharpening methods. We choose some representative methods as band-dependent spatial-detail with physical constraints approach (BDSD-PC) \cite{ref41}, generalized Laplacian pyramid with modulation transfer function-matched filters and a full scale regression-based injection model (MTF-GLP-FS) \cite{ref42}, a set of some competitive DL-based methods including PanNet \cite{ref11}, DiCNN \cite{ref44}, FusionNet \cite{ref19}, LAGConv \cite{ref16_1}, CTINN \cite{ref45}, LGPConv \cite{ref46_1} and MMNet \cite{ref46}. A recently proposed diffusion-based pansharpening method, PanDiff \cite{ref32}, CrossDiff \cite{ref34} and SSDiff \cite{ref33} are also compared. The above methods were implemented using DLPan-Toolbox \cite{ref39}.


\subsection{Evaluation Metrics}
To quantitatively evaluate the rationality and superiority of the proposed method, we introduce several evaluation metrics at both reduced- and full-resolution levels in our experiments. For reduced resolution, the metrics include peak signal-to-noise ratio (PSNR), structural similarity index (SSIM) \cite{ref47}, spectral angle mapper (SAM) \cite{ref48}, erreur relative globale adimensionnelle de synthèse (ERGAS) \cite{ref49}, and spatial correlation coefficient (SCC) \cite{ref50}. These metrics are effective in distinguishing various features of the fused images. Specifically, SSIM and SCC are more adept at measuring spatial similarity, while SAM is tailored for assessing spectral differences. PSNR and ERGAS provide a comprehensive evaluation by considering both spectral and spatial differences. For full-resolution experiments, since there is no ground truth available, we use non-reference metrics to validate the accuracy. These metrics include $D_\lambda$ and $D_S$ which assess spectral and spatial distortions respectively, and hybrid quality with no reference (HQNR) \cite{ref52}. The HQNR index has an ideal value of $1$, while $D_\lambda$ and $D_S$ have ideal values of $0$.

\section{Experimental Results}

\subsection{Results on WorldView-3 Dataset}

In this section, we present qualitative and quantitative results of the proposed method and comparisons on the WV3 dataset from PanCollection. Reduced-resolution experiments use $256 \times 256$ IMS, PAN, and ground truth images for visual clarity, while full-resolution experiments use $512 \times 512$ PAN and IMS images. Results for both resolutions are shown in Table \ref{tab:wv3}. Fusion results and error maps are displayed in Fig. 6, with a zoomed-in region for detailed analysis.

As shown in Tab. \ref{tab:wv3}, deep learning-based methods outperform traditional methods significantly. Among them, our approach achieves superior performance on both reduced- and full-resolution datasets. HFreqDiff excels in spatial enhancement and spectral fidelity, achieving the lowest SAM and highest SCC on the reduced-resolution dataset, and the highest HQNR on the full-resolution dataset.

\begin{figure*}[htbp]
\centering
\subcaptionbox*{}{\includegraphics[width=0.135\textwidth]{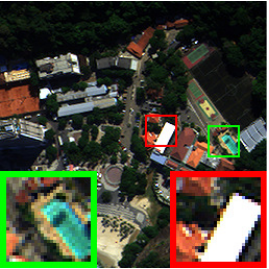}}
\subcaptionbox*{}{\includegraphics[width=0.135\textwidth]{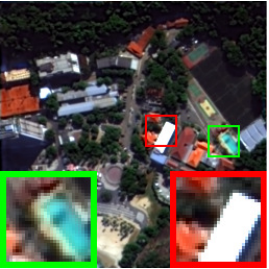}}
\subcaptionbox*{}{\includegraphics[width=0.135\textwidth]{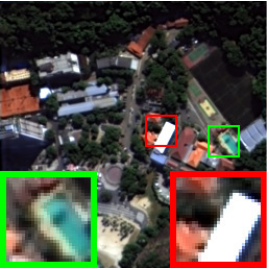}}
\subcaptionbox*{}{\includegraphics[width=0.135\textwidth]{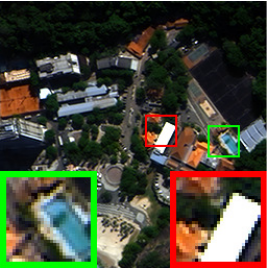}}
\subcaptionbox*{}{\includegraphics[width=0.135\textwidth]{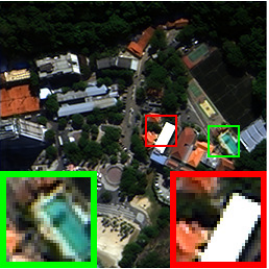}}
\subcaptionbox*{}{\includegraphics[width=0.135\textwidth]{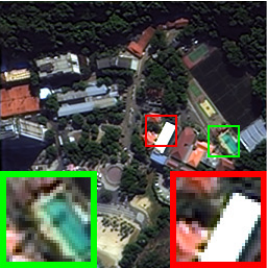}}
\subcaptionbox*{}{\includegraphics[width=0.135\textwidth]{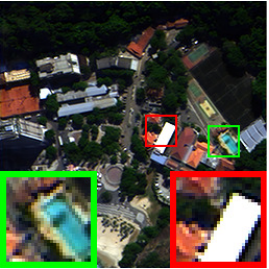}}
\\ \vspace{-16 pt}
\subcaptionbox*{ground truth}{\includegraphics[width=0.135\textwidth]{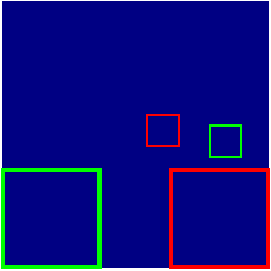}}
\subcaptionbox*{BDSD-PC}{\includegraphics[width=0.135\textwidth]{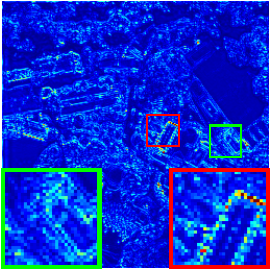}}
\subcaptionbox*{MTF-GLP-FS}{\includegraphics[width=0.135\textwidth]{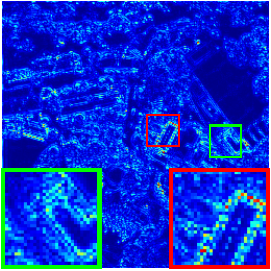}}
\subcaptionbox*{PanNet}{\includegraphics[width=0.135\textwidth]{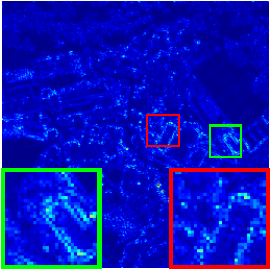}}
\subcaptionbox*{DiCNN}{\includegraphics[width=0.135\textwidth]{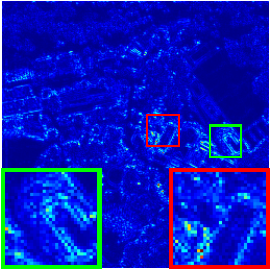}}
\subcaptionbox*{FusionNet}{\includegraphics[width=0.135\textwidth]{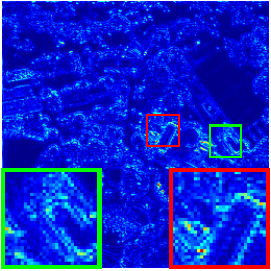}}
\subcaptionbox*{LAGConv}{\includegraphics[width=0.135\textwidth]{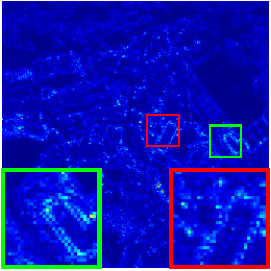}}
\\
\subcaptionbox*{}{\includegraphics[width=0.135\textwidth]{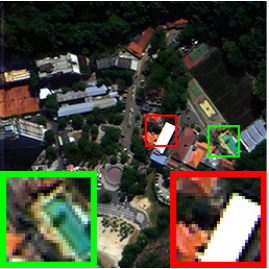}}
\subcaptionbox*{}{\includegraphics[width=0.135\textwidth]{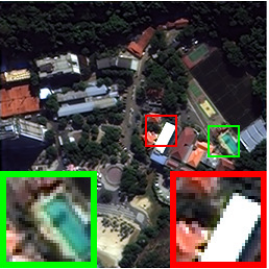}}
\subcaptionbox*{}{\includegraphics[width=0.135\textwidth]{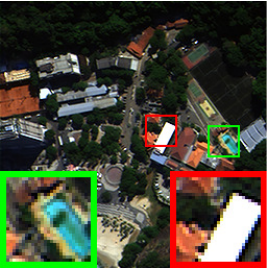}}
\subcaptionbox*{}{\includegraphics[width=0.135\textwidth]{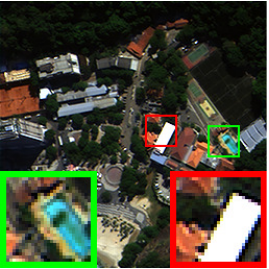}}
\subcaptionbox*{}{\includegraphics[width=0.135\textwidth]{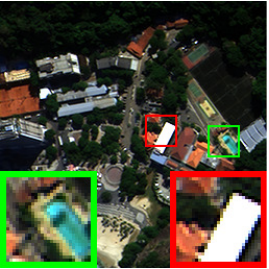}}
\subcaptionbox*{}{\includegraphics[width=0.135\textwidth]{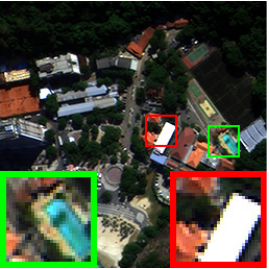}}
\subcaptionbox*{}{\includegraphics[width=0.135\textwidth]{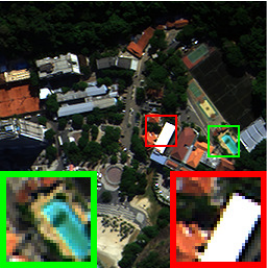}}
\\\vspace{-16 pt}  
\subcaptionbox*{CTINN}{\includegraphics[width=0.135\textwidth]{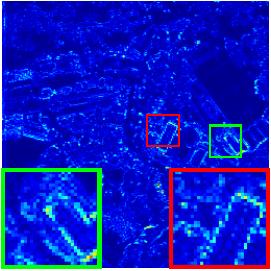}}
\subcaptionbox*{LGPConv}{\includegraphics[width=0.135\textwidth]{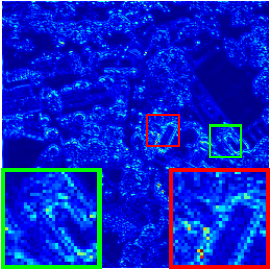}}
\subcaptionbox*{MMNet}{\includegraphics[width=0.135\textwidth]{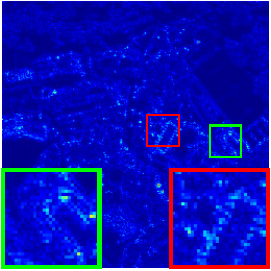}}
\subcaptionbox*{PanDiff}{\includegraphics[width=0.135\textwidth]{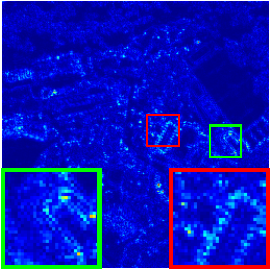}}
\subcaptionbox*{CrossDiff}{\includegraphics[width=0.135\textwidth]{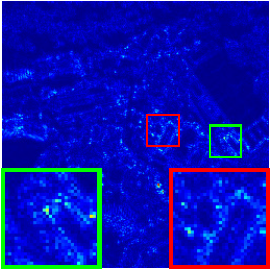}}
\subcaptionbox*{SSDiff}{\includegraphics[width=0.135\textwidth]{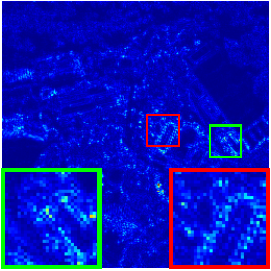}}
\subcaptionbox*{HFreqDiff (ours)}{\includegraphics[width=0.135\textwidth]{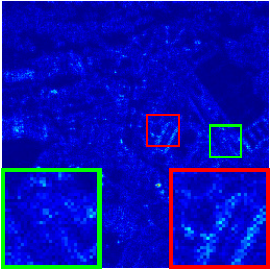}}
\caption{The fused images in the reduced-resolution test on WorldView-3 and the error maps between the fusion results and the ground truth.} \label{fig:wv3_red}
\end{figure*}

Fig. \ref{fig:wv3_red} displays the quantitative results of all methods on the reduced-resolution test. Error maps indicate that deep learning-based methods exhibit higher fusion accuracy, as shown by their darker error maps. HFreqDiff's fused MS image is notably closer to the ground truth, with the darkest error map. Local enlargements (red and green windows) further demonstrate that our method best preserves spatial texture while maintaining spectral consistency with the ground truth.




\begin{table*}[htbp]
	\centering
	\resizebox{.8\textwidth}{!}{
	\begin{tabular}{c|ccccccccc}
		\hline \hline
		\multirow{2}{*}{Method} & \multicolumn{5}{c}{Reduce}                                                             &  && Full                             &                 \\ \cline{2-10} 
		& PSNR           & SSIM            & SAM             & ERGAS           & SCC             &  & $D_{\lambda}$   & $D_s$           & HQNR             \\ \hline
		BDSD-PC \cite{ref41}                 & 34.16          & 0.8936          & 5.4270          & 4.2158          & 0.8896          &  & 0.1201          & 0.1164          & 0.7801          \\
		MTF-GLP-FS \cite{ref42}              & 34.34          & 0.8912          & 5.2064          & 4.1189          & 0.8887          &  & 0.0631          & 0.1390          & 0.8266          \\
		PanNet \cite{ref11}                 & 36.81          & 0.9617          & 3.7951          & 2.8347          & 0.9621          &  & 0.0862          & 0.0521          & 0.8921          \\
		DiCNN \cite{ref44}                  & 36.13          & 0.9442          & 4.3644          & 3.0155          & 0.9519          &  & 0.0460          & 0.0567          & 0.9174          \\
		FusionNet \cite{ref19}              & 36.42          & 0.9222          & 5.1869          & 3.8505          & 0.9516          &  & 0.0219          & 0.0435          & 0.9308          \\
		LAGConv \cite{ref16_1}                & 37.47          & 0.9787          & 3.6152          & 2.5999          & 0.9750          &  & 0.0386          & 0.0406          & 0.9401          \\
		CTINN \cite{ref45}                  & 38.32          & \textbf{0.9816} & 3.2523          & 2.3936          & \underline{0.9826}  &  & \underline{0.0215}          & 0.0309          & 0.9515          \\
		LGPConv \cite{ref46_1}                & 37.88          & 0.9709          & 3.2884          & 2.4276          & 0.9808          &  & 0.0262          & \underline{0.0233}          & \underline{0.9667}          \\
		MMNet \cite{ref46}                 & 37.46          & \underline{0.9803} & 3.5271          & 2.6335          & 0.9774          &  & 0.0225          & 0.0245  & 0.9589          \\
		PanDiff \cite{ref32}                & 36.58          & 0.9444          & 3.2191          & 2.8682          & 0.9602          &  & 0.0234          & 0.0374          & 0.9591          \\
		CrossDiff \cite{ref34}              & 38.17 & 0.9742       & 2.9309 & 2.1861 & 0.9806          &  & 0.0327 & 0.0289          & 0.9578          \\
		SSDiff \cite{ref33}                 & \underline{39.04}         & 0.9714          & \underline{2.9167}          & 2.1365          & 0.9815          &  & 0.0253          & 0.0284          & 0.9569          \\
		HFreqDiff (ours)            & \textbf{39.48} & 0.9770          & \textbf{2.7866} & \textbf{2.0615} & \textbf{0.9847} &  & \textbf{0.0198} & \textbf{0.0204} & \textbf{0.9697} \\ \hline
		Ideal value             & $+\infty $     & 1               & 0               & 0               & 1               &  & 0               & 0               & 1               \\ \hline \hline
	\end{tabular}}
\caption{Quantitative metrics for all the comparison methods on the WV3 reduced-resolution and full-resolution datasets. The best results are highlighted in bold, while the second-best results are underlined.}\label{tab:wv3}
\end{table*}

\begin{figure*}[htbp]
\centering
\subcaptionbox{PanNet}{\includegraphics[width=\textwidth]{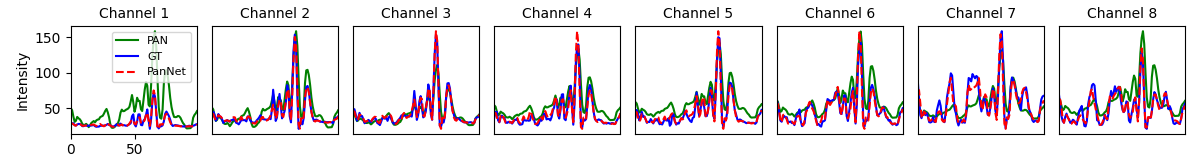}}
\subcaptionbox{PanDiff}{\includegraphics[width=\textwidth]{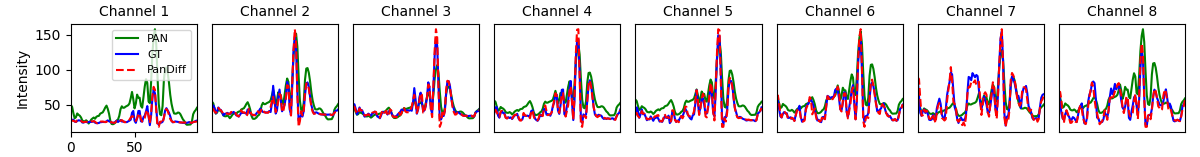}}
\subcaptionbox{HFreqDiff (ours)}{\includegraphics[width=\textwidth]{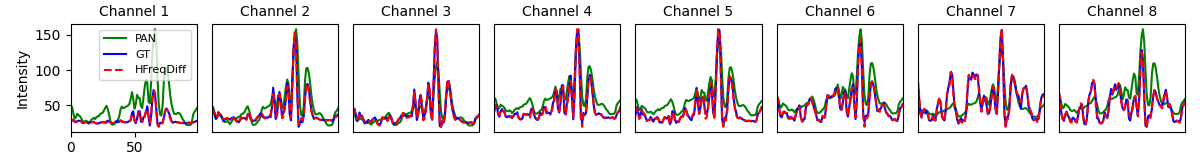}}
\caption{Scanning values of the ground truth image, PAN image and predicted results in WV3 reduced-resolution testing (transform to the range of $[0, 255]$). The x-axis represents pixels of the scanning line. (a) Results of PanNet; (b) Results of PanDiff (c) Results of HFreqDiff.} \label{fig:wv3_intensity}
\end{figure*}

Besides, the corresponding intensity curves along a randomly selected scanning line are presented in Fig. \ref{fig:wv3_intensity} for better visual comparison. We selected areas with significant changes in building clusters for visualization analysis to evaluate the model’s performance more intuitively. From the curves in the figures, it can be observed that the pixel intensity of certain channels in the MS image differs significantly from that of the PAN image, reflecting their intrinsic spectral differences. However, the pixel intensity curves generated by our model closely align with the ground truth, demonstrating its exceptional ability to capture and integrate multispectral information while accurately reconstructing the real pixel distribution.

\begin{figure*}[htbp]
\centering
\subcaptionbox*{PAN}{\includegraphics[width=0.135\textwidth]{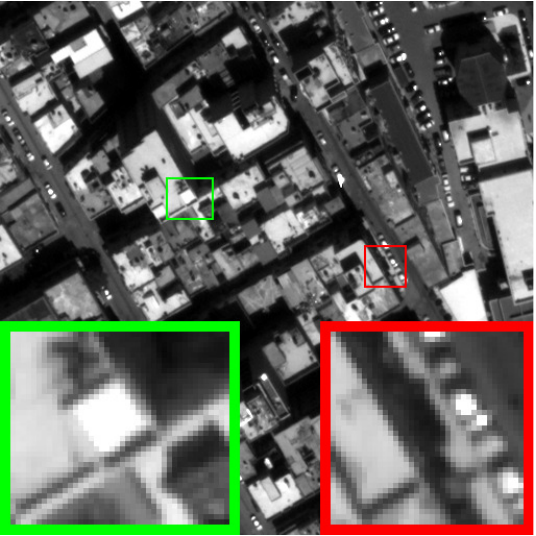}}
\subcaptionbox*{MTF-GLP-FS}{\includegraphics[width=0.135\textwidth]{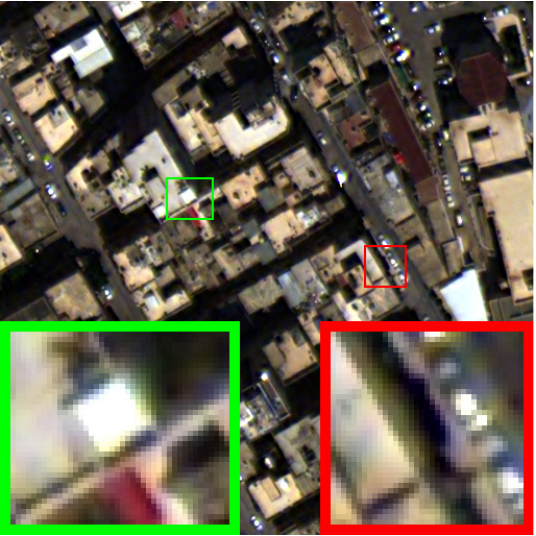}}
\subcaptionbox*{PanNet}{\includegraphics[width=0.135\textwidth]{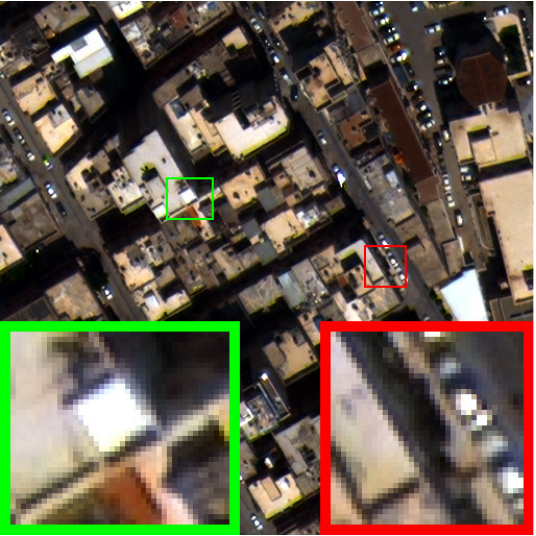}}
\subcaptionbox*{DiCNN}{\includegraphics[width=0.135\textwidth]{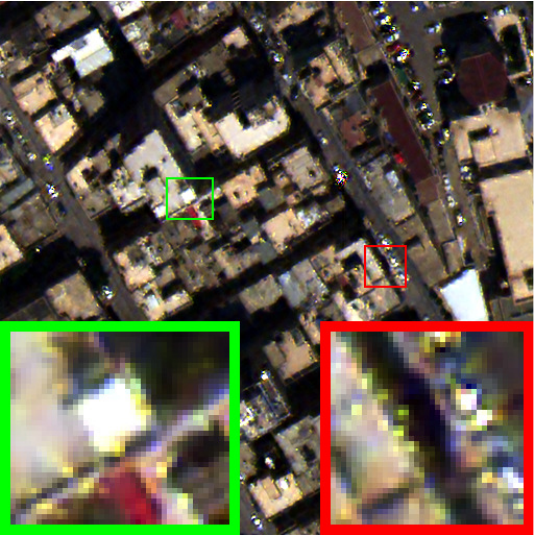}}
\subcaptionbox*{FusionNet}{\includegraphics[width=0.135\textwidth]{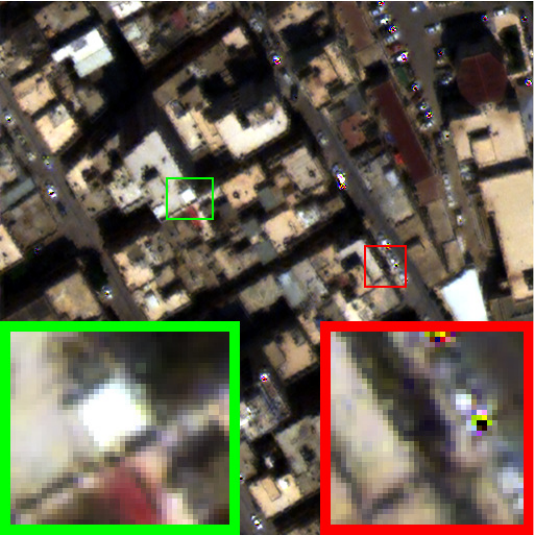}}
\subcaptionbox*{LAGConv}{\includegraphics[width=0.135\textwidth]{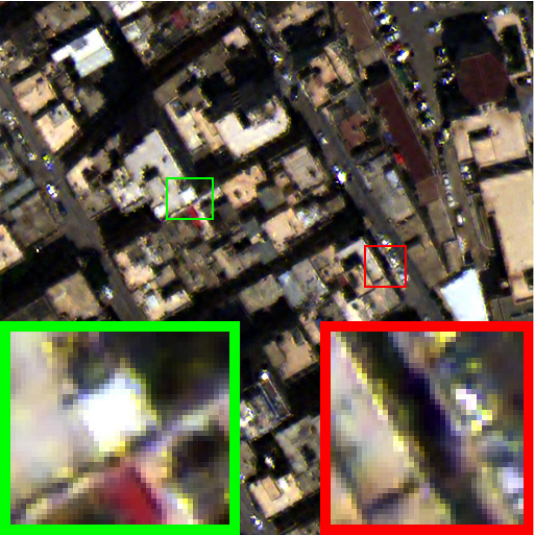}}
\subcaptionbox*{CTINN}{\includegraphics[width=0.135\textwidth]{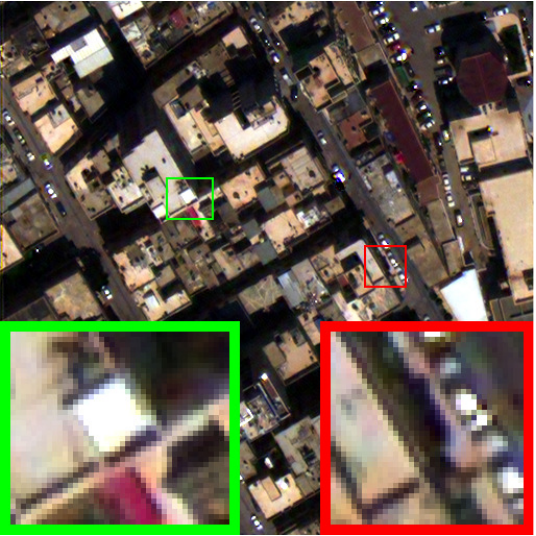}}
\\
\subcaptionbox*{IMS}{\includegraphics[width=0.135\textwidth]{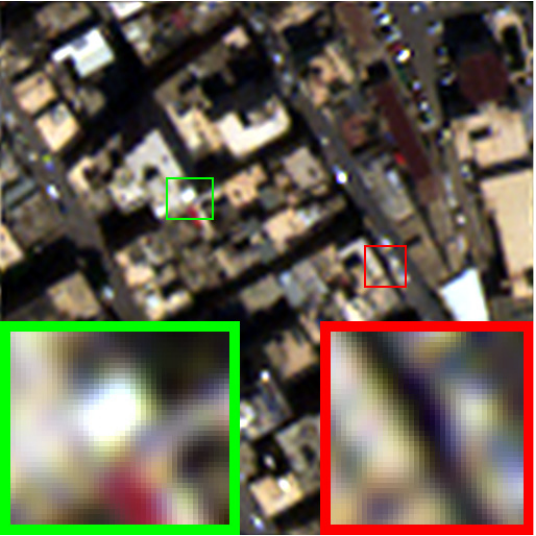}}
\subcaptionbox*{LGPConv}{\includegraphics[width=0.135\textwidth]{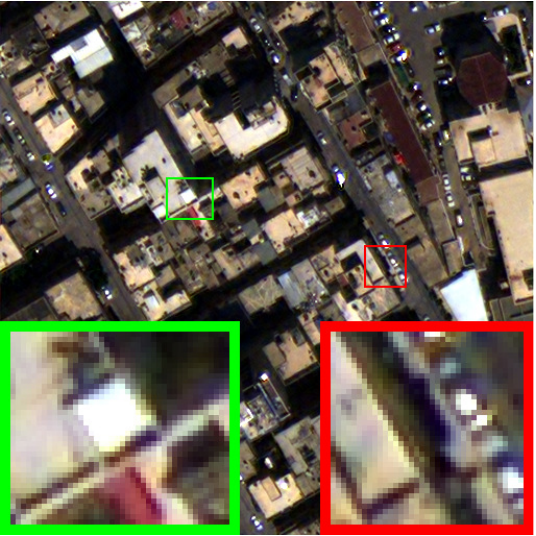}}
\subcaptionbox*{MMNet}{\includegraphics[width=0.135\textwidth]{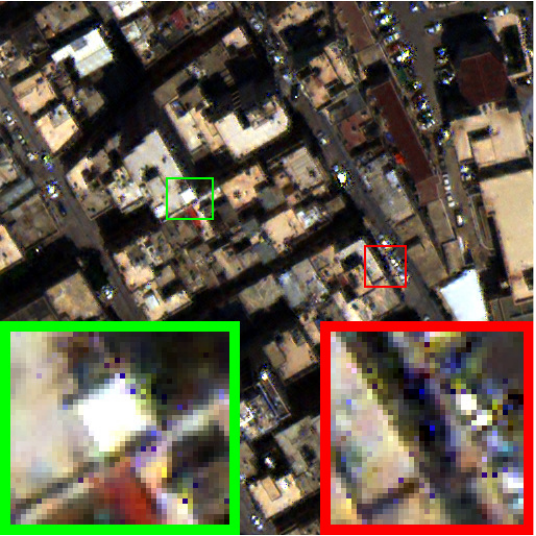}}
\subcaptionbox*{PanDiff}{\includegraphics[width=0.135\textwidth]{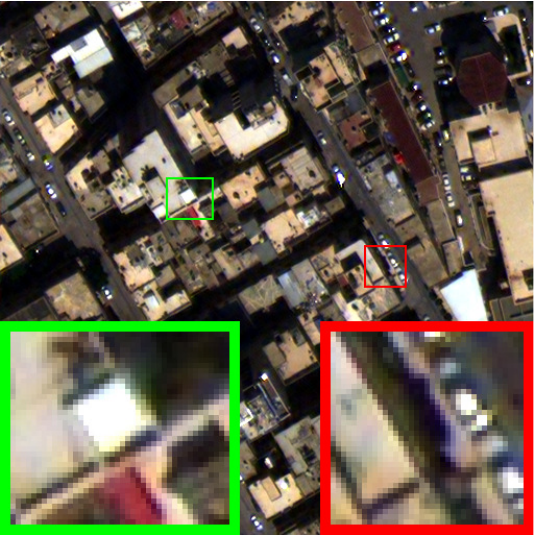}}
\subcaptionbox*{CrossDiff}{\includegraphics[width=0.135\textwidth]{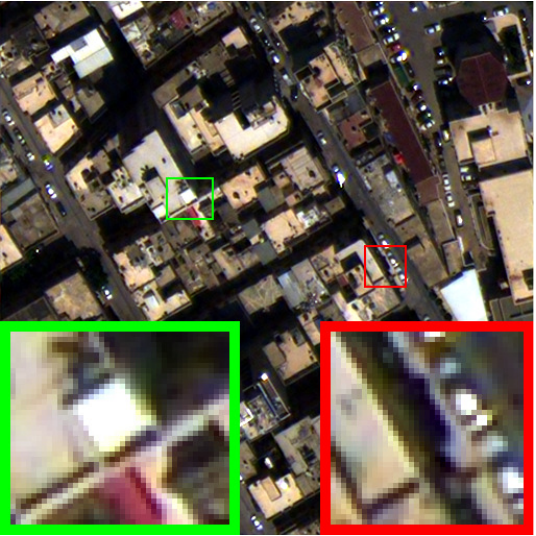}}
\subcaptionbox*{SSDiff}{\includegraphics[width=0.135\textwidth]{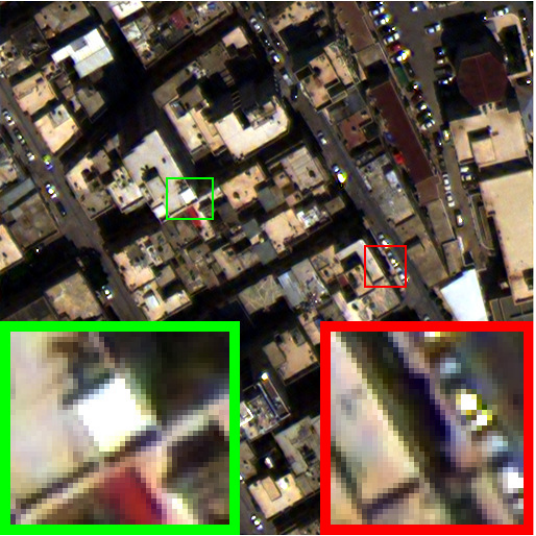}}
\subcaptionbox*{HFreqDiff (ours)}{\includegraphics[width=0.135\textwidth]{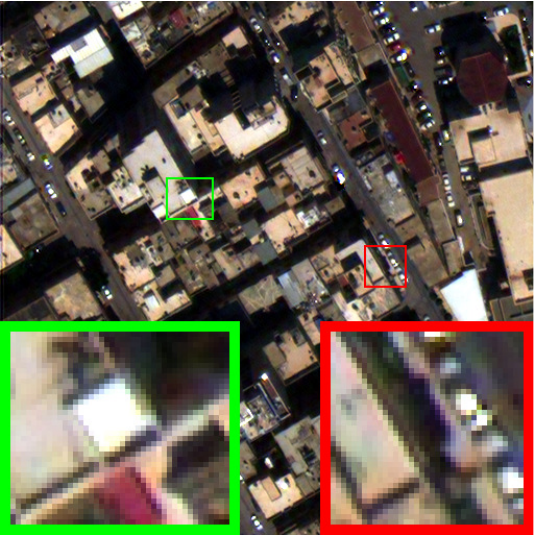}}
\caption{The fused images in the full-resolution test on WorldView-3 between the fusion results.} \label{fig:wv3_full}
\end{figure*}

The quantitative results of the full-resolution experiment are shown in Fig. \ref{fig:wv3_full}. HFreqDiff maintains its typical advantages over comparative methods. The obtained results indicate that our method can fuse HRMS images reducing spatial and spectral distortions, thus demonstrating that it has a good generalization at full-resolution.

\subsection{Results on the QuickBird dataset}


This section presents the test results on the QuickBird dataset. Table \ref{tab:qb} summarizes the results of all tested methods. Our approach outperforms the benchmark in almost all reference and non-reference metrics for both reduced-resolution and full-resolution datasets.

\begin{figure*}[htbp]
	\centering
    \subcaptionbox*{}{\includegraphics[width=0.135\textwidth]{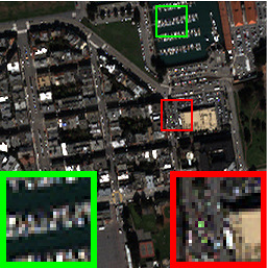}}
	\subcaptionbox*{}{\includegraphics[width=0.135\textwidth]{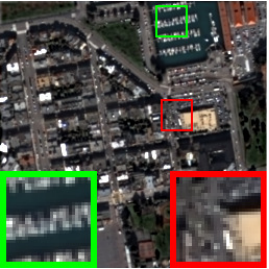}}
    \subcaptionbox*{}{\includegraphics[width=0.135\textwidth]{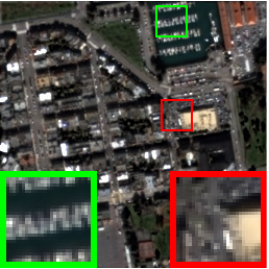}}
	\subcaptionbox*{}{\includegraphics[width=0.135\textwidth]{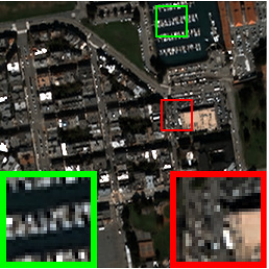}}
    \subcaptionbox*{}{\includegraphics[width=0.135\textwidth]{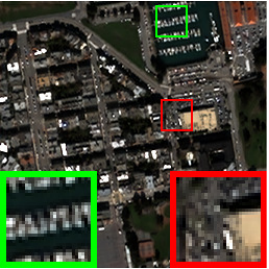}}
    \subcaptionbox*{}{\includegraphics[width=0.135\textwidth]{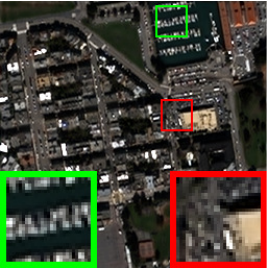}}
    \subcaptionbox*{}{\includegraphics[width=0.135\textwidth]{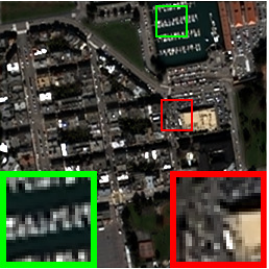}}
    \\\vspace{-16 pt}
    \subcaptionbox*{ground truth}{\includegraphics[width=0.135\textwidth]{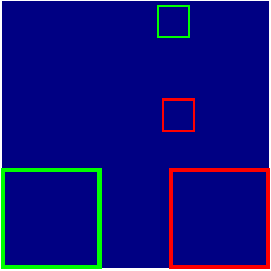}}
	\subcaptionbox*{BDSD-PC}{\includegraphics[width=0.135\textwidth]{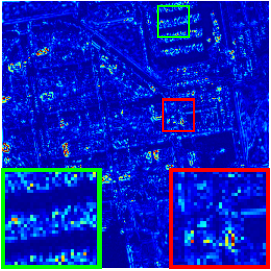}}
    \subcaptionbox*{MTF-GLP-FS}{\includegraphics[width=0.135\textwidth]{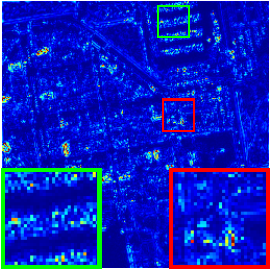}}
	\subcaptionbox*{PanNet}{\includegraphics[width=0.135\textwidth]{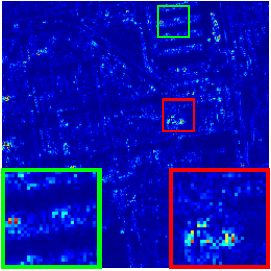}}
    \subcaptionbox*{DiCNN}{\includegraphics[width=0.135\textwidth]{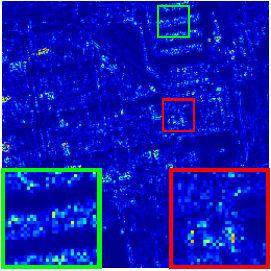}}
    \subcaptionbox*{FusionNet}{\includegraphics[width=0.135\textwidth]{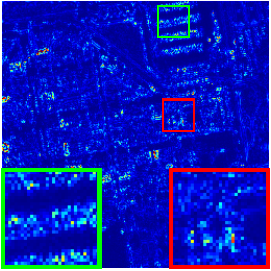}}
    \subcaptionbox*{LAGConv}{\includegraphics[width=0.135\textwidth]{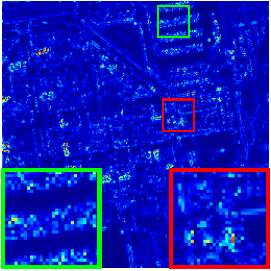}}
    \\
    \subcaptionbox*{}{\includegraphics[width=0.135\textwidth]{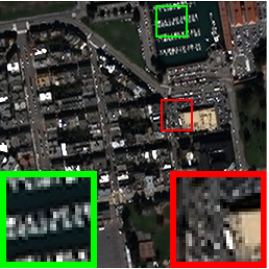}}
    \subcaptionbox*{}{\includegraphics[width=.135\textwidth]{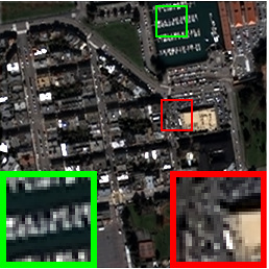}}
    \subcaptionbox*{}{\includegraphics[width=0.135\textwidth]{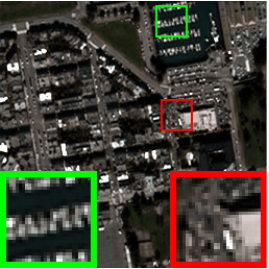}}
    \subcaptionbox*{}{\includegraphics[width=0.135\textwidth]{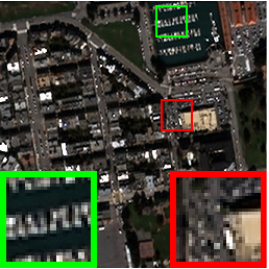}}
    \subcaptionbox*{}{\includegraphics[width=0.135\textwidth]{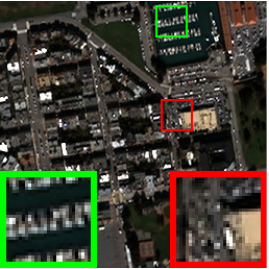}}
    \subcaptionbox*{}{\includegraphics[width=0.135\textwidth]{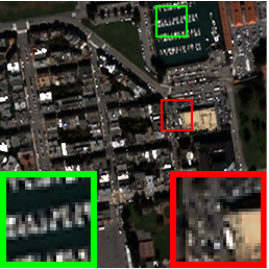}}
    \subcaptionbox*{}{\includegraphics[width=0.135\textwidth]{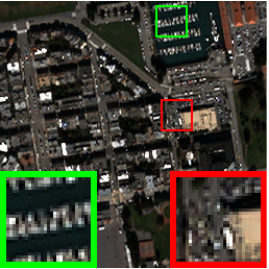}}
    \\\vspace{-16 pt}
    \subcaptionbox*{CTINN}{\includegraphics[width=0.135\textwidth]{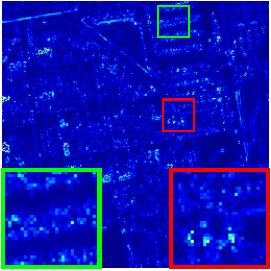}}
    \subcaptionbox*{LGPConv}{\includegraphics[width=.135\textwidth]{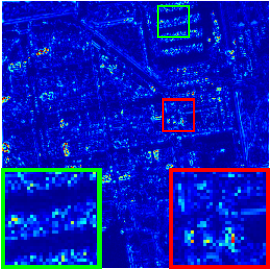}}
    \subcaptionbox*{MMNet}{\includegraphics[width=0.135\textwidth]{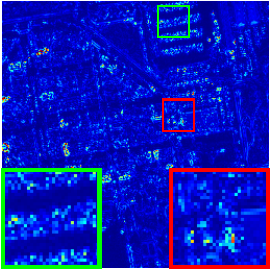}}
    \subcaptionbox*{PanDiff}{\includegraphics[width=0.135\textwidth]{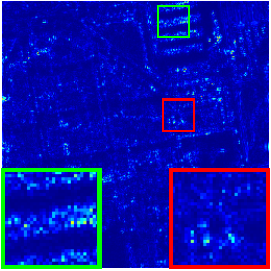}}
    \subcaptionbox*{CrossDiff}{\includegraphics[width=0.135\textwidth]{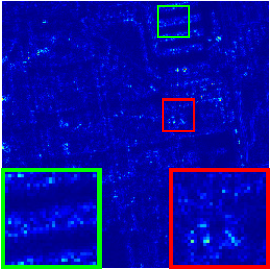}}
    \subcaptionbox*{SSDiff}{\includegraphics[width=0.135\textwidth]{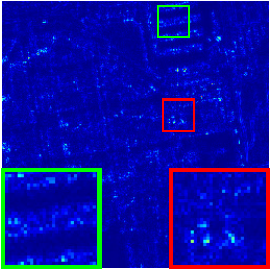}}
    \subcaptionbox*{HFreqDiff (ours)}{\includegraphics[width=0.135\textwidth]{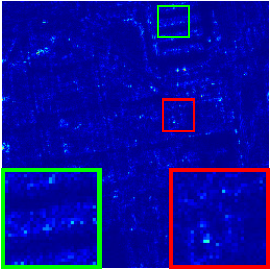}}
	\caption{The fused images in the reduced-resolution test on the QuickBird and the error maps between the fusion results and the ground truth.} \label{fig:qb_red}
\end{figure*}

From the results, it can be observed that HFreqDiff surpasses all competitors by a clear margin across all evaluation metrics at reduced resolution, particularly in the spectral-related indices SAM and ERGAS, where it achieves significant improvements over other models.
Additionally, as shown in Fig. \ref{fig:qb_red}, the colors and edge details produced by our method are closer to the ground truth. Traditional fusion methods exhibit higher brightness values in their error maps, whereas the error maps indicate that the residual between HFreqDiff and the ground truth is minimal.

\begin{table*}[htbp]
	\centering
	\resizebox{0.8\textwidth}{!}{
	\begin{tabular}{c|ccccccccc}
		\hline \hline
		\multirow{2}{*}{Method} & \multicolumn{5}{c}{Reduce}                                                             &  && Full                             &                 \\ \cline{2-10} 
		& PSNR           & SSIM            & SAM             & ERGAS           & SCC             &  & $D_{\lambda}$   & $D_s$           & HQNR             \\ \hline
		BDSD-PC \cite{ref41}                & 35.95          & 0.8957          & 6.7232          & 6.3606          & 0.8979          &  & 0.1345          & 0.1689          & 0.7406          \\
		MTF-GLP-FS \cite{ref42}             & 36.12          & 0.8967          & 6.4589          & 6.2240          & 0.8975          &  & 0.0674          & 0.1408          & 0.8134          \\
		PanNet \cite{ref11}                 & 35.57          & 0.9321          & 5.0604          & 5.2545          & 0.9532          &  & \underline{0.0475}          & 0.0462          & 0.8981          \\
		DiCNN \cite{ref44}                  & 35.41          & 0.9266          & 5.1528          & 5.4922          & 0.9523          &  & 0.0985          & 0.1081          & 0.8149          \\
		FusionNet \cite{ref19}              & 35.53          & 0.9325          & 5.5137          & 5.2579          & 0.9527          &  & 0.0599          & 0.0582          & 0.8852          \\
		LAGConv \cite{ref16_1}                & 34.39          & 0.9227          & 5.5473          & 5.1999          & 0.9503          &  & 0.0844          & 0.0676          & 0.8682          \\
		CTINN \cite{ref45}                  & 35.85          & 0.9375          & 5.7261          & 5.0203          & 0.9561          &  & 0.0506          & 0.0694          & 0.8390          \\
		LGPConv \cite{ref46_1}                & 35.77          & 0.9198          & 5.4987          & 4.7183          & 0.9491          &  & 0.0780          & 0.0530          & 0.8543          \\
		MMNet \cite{ref46}                 & 36.47          & 0.9250          & 5.0030          & 4.9201          & 0.9614          &  & \textbf{0.0417}          & 0.0996          & 0.8661          \\
		PanDiff \cite{ref32}                & 37.35          & 0.9479          & 4.9866          & 4.2257          & 0.9656          &  & 0.0493          & 0.0542          & 0.8911          \\
		CrossDiff \cite{ref34}              & 38.31          & 0.9562          & 4.5752          & 3.7823          & 0.9658          &  & 0.0721          & 0.0787          & 0.8701          \\
		SSDiff \cite{ref33}              & \underline{38.44}  &\underline{0.9575}& \underline{4.5396} & \underline{3.6786}  & \underline{0.9765}          &  & 0.0501          & \underline{0.0431}          & \underline{0.8996}        \\
		HFreqDiff (ours)             & \textbf{39.89} & \textbf{0.9652} & \textbf{4.3107} & \textbf{3.3501} & \textbf{0.9798} &  & 0.0485 & \textbf{0.0396} & \textbf{0.9010} \\ \hline
		Ideal value             & $+\infty $             & 1               & 0               & 0               & 1               &  & 0               & 0               & 1               \\ \hline \hline
	\end{tabular}}
\caption{Quantitative metrics for all the comparison methods on the QuickBird reduced-resolution and full-resolution datasets. The best results are highlighted in bold, while the second-best results are underlined.}\label{tab:qb}
\end{table*}

\subsection{Ablation Study}
\subsubsection{PADM vs. Wald's Protocol}
To validate the effectiveness of PADM, we conducted comparative experiments by training the network on MS data preprocessed using the Wald's protocol on three datasets. Table \ref{tab:ablation_padm} presents the comparison results. It is evident that applying PADM to process MS data significantly enhances performance.

\begin{table}[htbp]
	\centering
	\begin{tabular}{cc|ccccc}
		\hline \hline
		&      & PSNR           & SSIM            & SAM             & ERGAS           & SCC             \\ \hline
		\multicolumn{1}{c|}{\multirow{2}{*}{WV3}} & Wald & 39.08          & 0.9748          & 2.8711          & 2.1003          & 0.9831          \\
		\multicolumn{1}{c|}{}                     & PADM & \textbf{39.48} & \textbf{0.9770} & \textbf{2.7766} & \textbf{2.0415} & \textbf{0.9846} \\ \hline
		\multicolumn{1}{c|}{\multirow{2}{*}{QB}}  & Wald & 39.01          & 0.9579          & 4.5195          & 3.6768          & 0.9756          \\
		\multicolumn{1}{c|}{}                     & PADM & \textbf{39.89} & \textbf{0.9652} & \textbf{4.3107} & \textbf{3.3501} & \textbf{0.9798} \\ \hline \hline
	\end{tabular}
\caption{Ablation study about PADM and Wald's Protocol for the WV3 and QB dataset. The best results are highlighted in bold.}\label{tab:ablation_padm}
\end{table}

\begin{table}[htbp]
\begin{tabular}{ccccccc}
\hline \hline
\multicolumn{7}{c}{WV3 (Reduced-Resolution)}                                                                                                  \\ \hline
                         & \multicolumn{1}{c|}{}     & PSNR           & SSIM            & SAM             & ERGAS           & SCC             \\ \hline
\multirow{2}{*}{PanNet}  & \multicolumn{1}{c|}{Wald} & 36.81          & 0.9617          & 3.7951          & 2.8347          & 0.9721          \\
                         & \multicolumn{1}{c|}{PADM} & \textbf{36.97} & \textbf{0.9636} & \textbf{3.7554} & \textbf{2.8029} & \textbf{0.9741} \\ \cline{2-7} 
\multirow{2}{*}{LAGConv} & \multicolumn{1}{c|}{Wald} & 37.47          & 0.9787          & 3.6152          & 2.5999          & 0.9750          \\
                         & \multicolumn{1}{c|}{PADM} & \textbf{37.69} & \textbf{0.9790} & \textbf{3.5513} & \textbf{2.5104} & \textbf{0.9798} \\ \hline \hline
\end{tabular}
\caption{Ablation Study on CNN Performance with PADM and Wald's Protocol on the WV3 Dataset. The best results are highlighted in bold.}
\end{table}

To further validate the effectiveness of PADM, we conducted additional experiments on the PanNet\cite{ref11} and LAGConv\cite{ref16_1} networks using the WV3 dataset. We compared the performance of networks trained on MS data generated by PADM and by the Wald's protocol. Table 1 shows the result, which shows that using PADM enables the network to perform better across various metrics, confirming the potential of PADM in degraded MS data generation and network training. This demonstrates PADM’s ability to generate more realistic degraded MS data and significantly improve network performance.

\subsubsection{Effectiveness of the High-frequency Feature Condition}
To assess the contribution of HDLM, we trained the diffusion model under the following configurations to isolate the impact of the proposed modules on WV3 dataset: (1) using only the concatenation of conditions and input together along the channel dimension as the network input; (2) employing only the CFB to process $PAN$ and $IMS$ for conditional guidance; (3) utilizing only the BAMB module to process $P^h$ and $IMS$ for conditional guidance; and (4) Both the CFB and BAMB modules are applied for conditional guidance.

\begin{table}[htbp]
	\centering
	\begin{tabular}{cc|ccccc}
    \hline \hline
    \multicolumn{7}{c}{WV3 (Reduced-Resolution)}                             \\ \hline
    CFB         & BAMB         & PSNR           & SSIM            & SAM      & ERGAS           & SCC          \\ \hline
                              &                           & 38.83          & 0.9655          & 3.2334          & 2.4501          & 0.9717          \\
    $\checkmark$ &                           & 39.27          & 0.9706          & 2.8532          & 2.0986          & 0.9818          \\
                              & $\checkmark$ & 39.07          & 0.9699          & 2.9039          & 2.2503          & 0.9787          \\
    $\checkmark$ & $\checkmark$ & \textbf{39.48} & \textbf{0.9770} & \textbf{2.7866} & \textbf{2.0615} & \textbf{0.9846} \\ \hline \hline
    \end{tabular}
\caption{Ablation study about HDLM on WV3 dataset. The best results are highlighted in bold.}\label{tab:ablation_hdlm}
\end{table}

Table \ref{tab:ablation_hdlm} shows a comparison of different network architectures and information injection strategies. It can be observed that simply concatenating the conditional inputs and training the network results in suboptimal performance. The CFB module provides deep-level guidance for $PAN$ and $IMS$, thereby improving the network's performance. With the introduction of BAMB, the network receives more refined high-frequency information guidance, ultimately achieving the best performance.

\FloatBarrier

\section{ Conclusion}
In this paper, we introduce the Progressive Alignment Degradation Module (PADM), an unsupervised framework designed to approximate degradation processes. Unlike existing deep learning-based methods that rely on fixed degradation models, PADM employs an iterative alternating optimization scheme to learn degradation, addressing the spatial distribution discrepancies between simulated and real low-resolution data. This approach provides novel insights into degradation modeling for image fusion tasks and effectively tackles nonlinear degradation challenges in complex scenarios. Building on PADM, we propose HFreqDiff, a diffusion model enhanced with a high-frequency feature learning module to extract critical details from PAN images and compensate for missing information in low-resolution images. Two feature modulation modules, CFB and BAMB, are incorporated to provide precise guidance, leveraging generative models to produce high-resolution multispectral (HRMS) images. Experiments on widely used pansharpening datasets demonstrate that our method outperforms state-of-the-art approaches both qualitatively and quantitatively. Future work will focus on integrating conditional information into the diffusion process for more accurate HRMS generation, improving degradation simulation for better generalization, and developing unsupervised models to reduce dependency on labeled data, thereby enhancing adaptability and robustness across diverse applications.


\section*{Acknowledgements}
The authors would like to thank the support by National Natural Science Foundation of China (12371419), Natural Sciences Foundation of Heilongjiang Province (ZD2022A001) and the National Natural Science Foundation of China (12271130, U21B2075, 12171123, 12401557).

\vfill

\end{document}